%% file: main.tex
\documentclass[10pt,twocolumn,letterpaper]{article}

\usepackage[pagenumbers]{iccv} 

\input{preamble}

\definecolor{iccvblue}{rgb}{0.21,0.49,0.74}
\usepackage[pagebackref,breaklinks,colorlinks,allcolors=iccvblue]{hyperref}

\def\paperID{6575} 
\def\confName{ICCV}
\def\confYear{2025}

\title{Joint Diffusion Models in Continual Learning}

\author{\textbf{Paweł Skierś}\\
Warsaw University of Technology\\
{\tt\small pawel.skiers.stud@pw.edu.pl}
\and
\textbf{Kamil Deja}\\
Research Institute IDEAS\\
Warsaw University of Technology\\
{\tt\small kamil.deja@pw.edu.pl}
}

\begin{document}
\maketitle

\begin{abstract}
In this work, we introduce \ours{} -- a new method for continual learning with generative rehearsal based on joint diffusion models. Neural networks suffer from catastrophic forgetting defined as an abrupt loss in the model's performance when retrained with additional data coming from a different distribution. Generative-replay-based continual learning methods try to mitigate this issue by retraining a model with a combination of new and rehearsal data sampled from a generative model. In this work, we propose to extend this idea by combining a continually trained classifier with a diffusion-based generative model into a single -- jointly optimized neural network. We show that such shared parametrization, combined with the knowledge distillation technique, allows for stable adaptation to new tasks without catastrophic forgetting. We evaluate our approach on several benchmarks, where it outperforms recent state-of-the-art generative replay techniques. Additionally, we extend our method to the semi-supervised continual learning setup, where it outperforms competing buffer-based replay techniques, and evaluate, in a self-supervised manner, the quality of trained representations. 
\end{abstract}
\vspace{-0.8cm}
\section{Introduction}
\vspace{-0.2cm}
Contemporary deep neural networks can be trained to human-level performance on a variety of different problems. However, contrary to humans, when retraining the same models to learn an additional task, neural networks suffer from \emph{catastrophic forgetting}~\cite{1999french} defined as an abrupt loss in performance on the previous task when adapting to a new one. Continual learning approaches try to mitigate this issue. In particular, replay-based techniques retrain the model with a mix of new data examples and samples from the past tasks stored in the memory buffer~\cite{2019rolnick+4,aljundi2019online}. However, since the memory buffer has to constantly grow, new methods~\cite{2017shin+3, van2020brain} use generative models to generate synthetic versions of past experiences. 

In the majority of the recent works, the generative replay technique follows the simple algorithm: 

\begin{algorithm}[H]
\small
\caption{Generative replay}
\begin{algorithmic}
    \State \textbf{Input:} $G$, $C$, sequence of tasks $\{D_i\}_{i=1}^N$
    \State $G \gets$ \textbf{train\_generative\_model}($G, D_1$)
    \State $C \gets$ \textbf{train\_classifier}($C, D_1$)
    \For{i in 2 \dots $N$}
        \State $S \gets$ \textbf{sample\_dataset}($G$)
        \State $B \gets D_i \cup  S$
        \State $G \gets$ \textbf{train\_generative\_model}($G$, $B$)
        \State $C \gets$ \textbf{train\_classifier}($C$, $B$)
    \EndFor
\end{algorithmic}
\end{algorithm}
\vspace{-0.4cm}
Such an approach introduces an imbalance between the generative model and the classifier. The generative model plays a crucial role as a knowledge consolidation method that stores information about old tasks, while the classifier is highly dependent on the quality of generated samples, and the effectiveness of the knowledge transfer from the generative part of the system. Consequently, several studies~\cite{deja2021binplay, wu2018memory, lesort2019generative} suggest that it may be advantageous to reset the classifier's weights entirely and retrain it from scratch for each task, using a blend of new data and sampled generations -- a solution that contradicts the principles of continual learning.

At the same time, stable classifier retraining using only generated data remains a challenge. 
Even state-of-the-art diffusion-based generative models are known to have significant limitations in precise modeling of data distribution~\cite{sehwag2022generating}, which results in the degradation of the classifier's performance. 

In this work, we propose to address this limitation by introducing a joint model that combines the generative and discriminative models into a single parametrization. In particular, we extract the latent features from the UNet~\cite{ronneberger2015unet} trained as a denoising diffusion model and use them for classification. We optimize our joint model with a sum of generative and discriminative objectives.
Such an approach directly removes the need for transferring the knowledge from the generative model to the classifier when retraining the model on a sequence of tasks. 

\begin{figure}
    \centering
    \includegraphics[width=\linewidth]{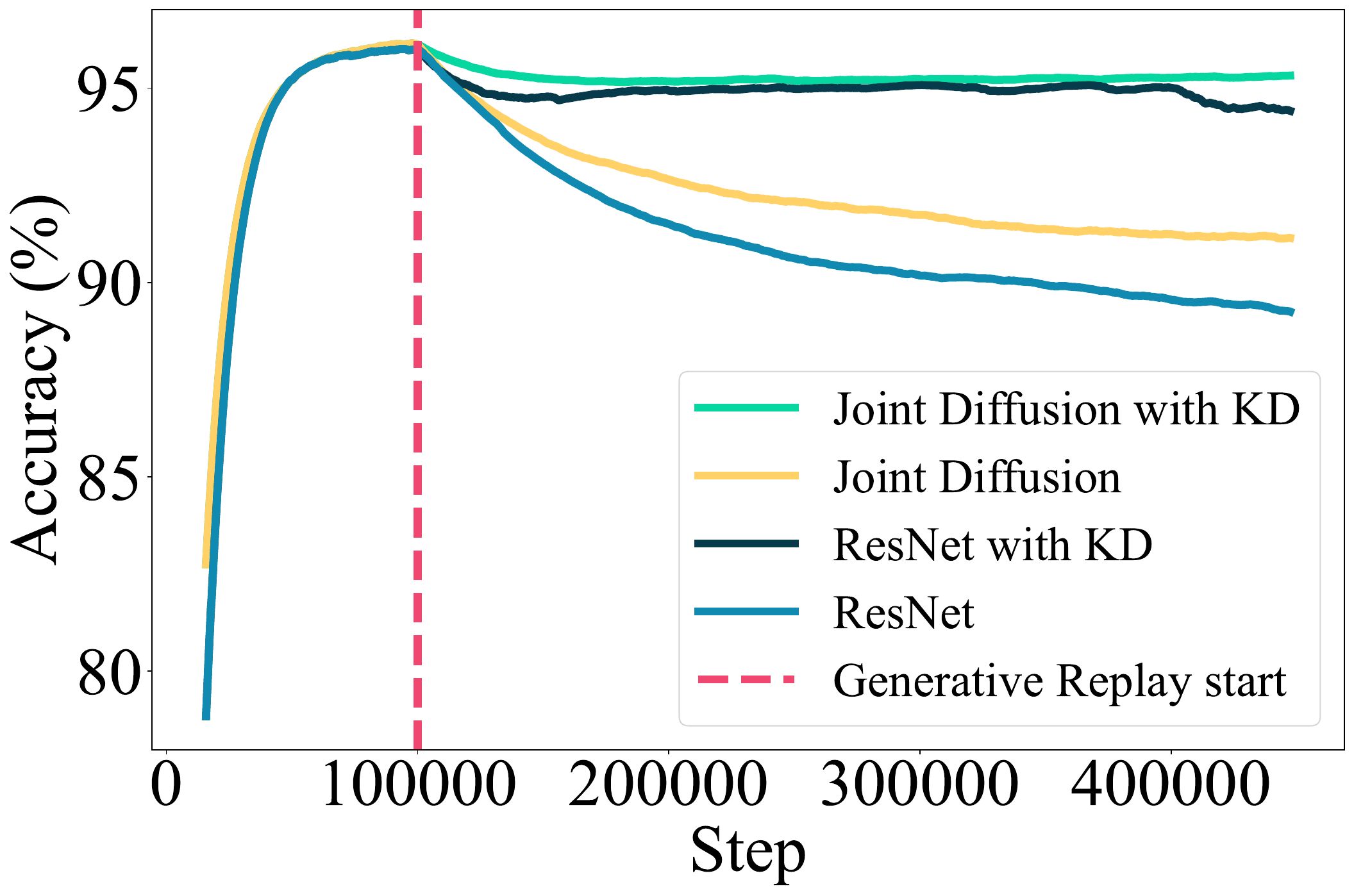}
    \vskip -0.2cm
    \caption{Further training of a classifier using data generated from the diffusion model trained with the same dataset harms the classifier's performance (blue line). However, joint modeling, especially in combination with knowledge distillation, significantly limits this degradation and retains almost the initial performance. We report accuracy averaged across three different seeds. }
    \vspace{-0.5cm}
    \label{fig:motivation_experiment}     
\end{figure}


To visualize the main benefit of this approach, we conduct a simple experiment that highlights the drawbacks of recent techniques. We train the classifier and our joint-diffusion model (ensuring that both models are of similar sizes) on the CIFAR10 dataset until convergence (for 100,000 optimizer steps)\footnote{Details on this experiment are provided in the Appendix}. Then, we continue to optimize the models, but instead of using real data, we sample data from the diffusion model. As presented in Fig.~\ref{fig:motivation_experiment}, the performance of the classifier (blue line), drastically decreases as the effect of the rehearsal with generated samples. At the same time, the performance of our joint diffusion (yellow) converges after a much smaller drop in performance.
Furthermore, the addition of knowledge distillation significantly decreases the degradation of both models, while keeping the trend of the joint diffusion model outperforming the classifier in terms of final performance and stability. 
We conclude that a combination of joint modeling and knowledge distillation is key for knowledge retention in generative replay-base CL. 


Therefore, we introduce a joint-diffusion continual learning method, that in combination with knowledge distillation and two-stage training allows for high-quality modeling in CL scenarios. In our experiments, we show that \ours{} achieves state-of-the-art performance when compared with other generative replay methods, while significantly reducing the computational cost. We show that this performance gain is attributed to the higher quality of rehearsal samples when compared with the standard diffusion model. Moreover, we show that our approach can be easily applied in the continual semi-supervised setup, where thanks to the shared parametrization we can observe an alignment between labeled and unlabeled data. In this task, we show the superiority of our method over the buffer-based techniques. Finally, we also highlight the quality and stability of representations built by our method by evaluating it in a self-supervised manner.
Our main contributions can be summarized as follows:
\begin{itemize}
    \item We introduce \ours{} -- a joint diffusion continual learning approach that mitigates the problem of knowledge transfer between generative and discriminative models in generative replay.
    \item We show that in class-incremental setup, our approach outperforms recent state-of-the-art generative replay techniques and approaches the soft-upper bound defined by a method with an infinite memory buffer, while significantly reducing the computational cost.
    \item We extend our studies to continual representation and semi-supervised learning where we compare \ours{} against buffer-based rehearsal techniques.
\end{itemize}

\section{Related Work}
\paragraph{Continual Learning}
The increasing size of machine learning models, and hence the increased computational cost of their training highlights the need for effective methods for model updates. Therefore, we can observe a growing number of continual learning techniques~\cite{parisi2019continual,mundt2023wholistic} that aim to mitigate the problem of catastrophic forgetting~\cite{1999french}, without reducing the ability to adjust to new data. CL methods can be categorized into three groups: (1) Regularization-based approaches such as EWC~\cite{kirkpatrick2017overcoming}, LwF~\cite{li2017learning} reduce the interference from the new tasks on the parts of the neural network that are crucial for the previous ones. (2) Architecture-based methods (e.g.RCL~\cite{2018xu+1}, PNN~\cite{2016rusu+7}) directly build different model versions for different tasks. Finally (3) Replay techniques (e.g.~\cite{2019rolnick+4,aljundi2019online}) use a limited number of past samples as \emph{rehearsal} examples, by mixing them with new data samples. 
\vspace{-0.4cm}
\paragraph{Generative models in continual learning}
A significant drawback of replay approaches is that storing past samples requires a growing memory buffer. Therefore, Shin et al.~\cite{2017shin+3} proposed to replace it with a Generative Adversarial Network~\cite{goodfellow2014generative}. This idea was further extended to VAEs~\cite{van2018generative,2020mundt+4}, GMMs~\cite{rostami2019complementary}, or Diffusion-based generative models~\cite{gao2023ddgr, cywiński2024guide}. In~\cite{lesort2019generative} authors provide a summary of different generative replay techniques. Apart from different architectures, similar to our idea, an interesting approach is proposed in Replay Through Feedback (RTF)~\cite{van2018generative}, where authors combine the variational autoencoder with a classifier. 
This improves the performance of the final solution, which is in line with results presented in~\cite{thai2021does, masarczyk2021onrob}, where authors show that additional reconstruction or generative loss added to the classifier alleviates catastrophic forgetting.

Several works further extend the basic generative replay technique. In~\cite{van2020brain}, followed by~\cite{liu2020generative}, authors propose to replay internal data representations instead of the original inputs. This approach improves model stability but limits its adaptation to new tasks. On the other hand, several methods drew inspiration from neuroscience and introduced techniques that mimic the human memory system by dividing the update of the model into two~\cite{kamra2017deep, deja2022multiband} or even three stages~\cite{wang2021triple}.

Recently, diffusion-based generative models (DDGMs)~\cite{sohl2015deep,ho2020denoising} have come to the forefront as a state-of-the-art generative method. Therefore, there is growing research on continual learning with DDGMs. Zajac et al.~\cite{zajac2023exploring} provide an overview of how DDGMs work when combined with baseline CL methods. In~\cite{gao2023ddgr} authors employ DDGMs with classifier guidance~\cite{dhariwal2021diffusion} as a source of rehearsal examples. This idea is further extended in~\cite{cywiński2024guide}, where authors overview different sampling strategies with DDGMs. There are several methods that explore the possibilities for efficient updates of large-scale text-to-image diffusion models. In DreamBooth~\cite{ruiz2023dreambooth} authors introduce a method for finetuning text-to-image DDGMs with \emph{Prior Preservation Loss}, that employs generative replay to mitigate catastrophic forgetting. Similarly, \cite{smith2023continual} extend this idea with Low Rank Adaptation (LORA~\cite{hu2021lora}) technique, while in~\cite{shah2023ziplora} apply this method for object and style update. On the other hand, Basu et al.~\cite{basu2023localizing}, show that knowledge in text-to-image models is highly localized and can be updated directly in the attention layers of the text encoder.
\vspace{-0.4cm}
\paragraph{H-space in diffusion models}

Several methods explore the potential of using the latent representations within the U-Net model used as a denoising decoder in DDGM. 
In particular, Kwon et al.~\cite{kwon2022diffusion} show that the bottleneck of the U-Net dubbed \emph{H-space} can be used as a semantic latent space for image manipulation. This idea is further extended in~\cite{park2023understanding}. Similarly, some techniques employ the H-space in downstream tasks outside of generative modeling. Authors of~\cite{baranchuk2021labelefficient}, \cite{tumanyan2023plug} and~\cite{rosnati2023robust}, show that such features can be used for image segmentation, in~\cite{luo2024diffusion} Luo et al. extend this idea to image correspondence, while ~\cite{deja2023learning} use H-features for classification. In this work, we relate to the utilization of H-space by employing it in a continually trained joint diffusion model with a classifier.
\vspace{-0.4cm}
\paragraph{Semi-supervised learning} 
Semi-supervised methods confront the problem of training models on large datasets where only a few samples have associated annotations \cite{fixmatch,surveyssl}. Recently, most of the best-performing semi-supervised methods belong to the family of hybrid methods that combine consistency regularization and pseudo-labeling ~\cite{mixmatch,remixmatch,fixmatch,surveyssl,flexmatch}. The main idea behind consistency regularization is an assumption that the model should yield similar outputs when fed perturbed versions of the same image or when its weights are slightly perturbed~\cite{laine2017temporal,regwithsochtransfandpertssl,meanteacher,virtadvregssl}. 
On the other hand, pseudo-labeling techniques exploit the concept of using synthetic labels for unlabeled data. More precisely, this involves utilizing "hard" labels (i.e., the arg max of the model's output) for which the predicted class probability surpasses a predetermined threshold~\cite{xie2020self,chen2020big,wang2020enaet,pham2021meta,lee2013pseudo,surveyssl}. 
Apart from consistency regularization and pseudo-labeling, a third group of SSL methods employs additional deep generative models ~\cite{surveyssl} 
such as Variational Auto-Encoder (VAE) \cite{li2019disentangled,paige2017learning,ehsan2017infinite,maaloe2016auxiliary,kingma2014semi}, or Generative Adversarial Networks (GAN) ~\cite{salimans2016improved,springenberg2015unsupervised,enhancedtgan,trianglegan,r3cgan}, to learn the underlying class-conditional distribution of partially labeled data. 

Despite their proliferation, as noted by~\cite{kang2022soft}, the standard semi-supervised approaches do not address the problem of learning in a continual setup. Consequently, a new line of research called continual semi-supervised learning emerged to address the challenges of the combination of those two domains. To that end,~\cite{wang2021ordisco} introduced ORDisCo, which utilizes a conditional GAN with a classifier to learn from partially labeled data in an incremental setup. On the other hand,~\cite{boschini2022continual} proposed CCIC, a method that enforces consistency by contrasting samples among different classes and tasks. In contrast, the recent state-of-the-art method called NNCSL~\cite{kang2022soft} leverages the nearest-neighbor classifier to learn the current task, while also retaining the relevant information from the previous tasks. 
In this work, we adapt the mechanism of pseudo-labeling and consistency regularization, which we combine with our joint diffusion model to combat the problems of semi-supervised continual learning.

\section{Background}

\subsection{Continual Learning}
In this work, we tackle the problem of continual learning, which as a machine learning paradigm assumes that data used for training of a function $f$ is provided in separate portions known as tasks. Particularly, in class-incremental learning~\cite{van2019three}, we assume that each task has a disjoint set of non-overlapping labels. In this setup, the objective is to minimize the prediction error on all of the labels without any additional information such as the identity of the associated task.

\subsection{Joint Diffusion Model}
\label{sec:jd}
        We follow the parametrization of joint diffusion models introduced in~\cite{deja2023learning}, where features from the different levels of the UNet architecture are pooled into a single vector $z$ used for classification.
        Let $p_\theta$ be the denoising model, which is of the UNet architecture, with parameters $\theta$. As such, the denoising model consists of the encoder $e_\nu$ and the decoder $d_\psi$ with parameters $\nu$ and $\psi$ respectively, such that $\theta = \{\nu,\psi\}$. The encoder takes input $x_t$ and returns a set of tensors $\mathcal{Z}_t= e_\nu(x_t) $ that is, a set of representation tensors $\mathcal{Z}_t = \{z_t^1, z_t^2 \dots z_t^n \}$ obtained from each depth level of the UNet architecture. The decoder reconstructs the denoised sample $x_{t-1}$, from the set of representations, i.e., $x_{t-1} = d_\psi(\mathcal{Z}_t)$. 
        Additionally to the standard diffusion denoiser, we propose to pool 
        the representation set into a single vector $z_t = f(\mathcal{Z}_t)$ via average pooling. Finally, we introduce classifier $g_\omega$, with parameters $\omega$ - the final part of the model that predicts target class $\hat{y} = g_\omega(z_t )$.
        
        
        We optimize the whole model jointly by modeling the joint probability as follows:
        \begin{equation}\label{eq:joint}
            p_{\nu, \psi, \omega}(x_{0:T}, y) = p_{\nu, \omega}(y | x_0)\ p_{\nu, \psi}(x_{0:T}),
        \end{equation}
        which after applying the logarithm yields:
        \begin{equation}\label{eq:joint_log}
            \ln p_{\nu, \psi, \omega}(x_{0:T}, y) = \ln p_{\nu, \omega}(y | x_0) + \ln p_{\nu, \psi}(x_{0:T}).
        \end{equation}
        We use the logarithm of the joint distribution (\ref{eq:joint_log}) as the training objective, where $\ln p_{\theta}(x_{0:T})$ is approximated by the simplified DDGM objective~\cite{huang2022improving}:
        \begin{equation}\label{eq:l_t_simple_ours}
            L_{t, \text{diff}}(\nu, \psi) = \mathbb{E}_{\mathbf{x}_{0}, \boldsymbol{\epsilon}}\left[\left\|\boldsymbol{\epsilon}-\boldsymbol{\hat{\epsilon}}\right\|^{2}\right],
        \end{equation}
        where $\epsilon$ is the noise added to the initial image and $\hat{\epsilon}$ is a prediction from the decoder:
        \begin{align}
             \{ z_t^1, z_t^2 \dots z_t^n \} =& e_\nu\left(\sqrt{\overline{\alpha}_{t}} \mathbf{x}_{0}+\sqrt{1-\overline{\alpha}_{t}} \boldsymbol{\epsilon}, t\right) \\
             \boldsymbol{\hat{\epsilon}} =& d_\psi(\{ z_t^1, z_t^2 \dots z_t^n \}) .
        \end{align}
        For the classifier, we employ the logarithm of the categorical distribution:
        \begin{equation}
            L_\text{class}(\nu, \omega) = -\mathbb{E}_{\mathbf{x}_0, y} \left[ \sum_{k=0}^{K-1} \mathds{1}[y=k] \log \frac{\exp \left( \varphi_{k} \right)}{\sum_{c=0}^{K-1} \exp \left( \varphi_{c} \right)} \right]
        \end{equation}
        which corresponds to the cross-entropy loss, where $y$ represents the target class, $\varphi$ is a vector of probabilities generated by the classifier $g_{\omega}(e_{\nu}(x_{0}))$ and $\mathds{1}[y=k]$ is the indicator function that is $1$ if $y$ equals $k$ and $0$ otherwise. The final loss function that we optimize with a single optimizer over parameters $\{ \nu, \psi, \omega \}$ is:
        \begin{align}
        \label{eq:final_joint}
            L_{JD}(\nu, \psi, \omega) = & \alpha \cdot L_\text{class}(\nu, \omega) - L_0(\nu, \psi) \notag\\ 
            & - \sum_{t=2}^{T} L_{t, \text{diff}}(\nu, \psi) - L_T(\nu, \psi) , 
        \end{align}
    
        To train the model over a batch of data examples $(x_0, y)$, we first noise the example $x_0$ with a forward diffusion to a random timestep $x_t$, so that the training loss for the denoising model is a Monte-Carlo approximation of the sum over all timesteps. After that, we feed $x_0$ to a classifier and calculate the cross-entropy loss on returned probabilities $\varphi$ and the given class labels $y$.
        

\section{Method}
        \begin{figure*}[t]
            \vskip -4mm
            \centering
                \includegraphics[width=.9\linewidth]{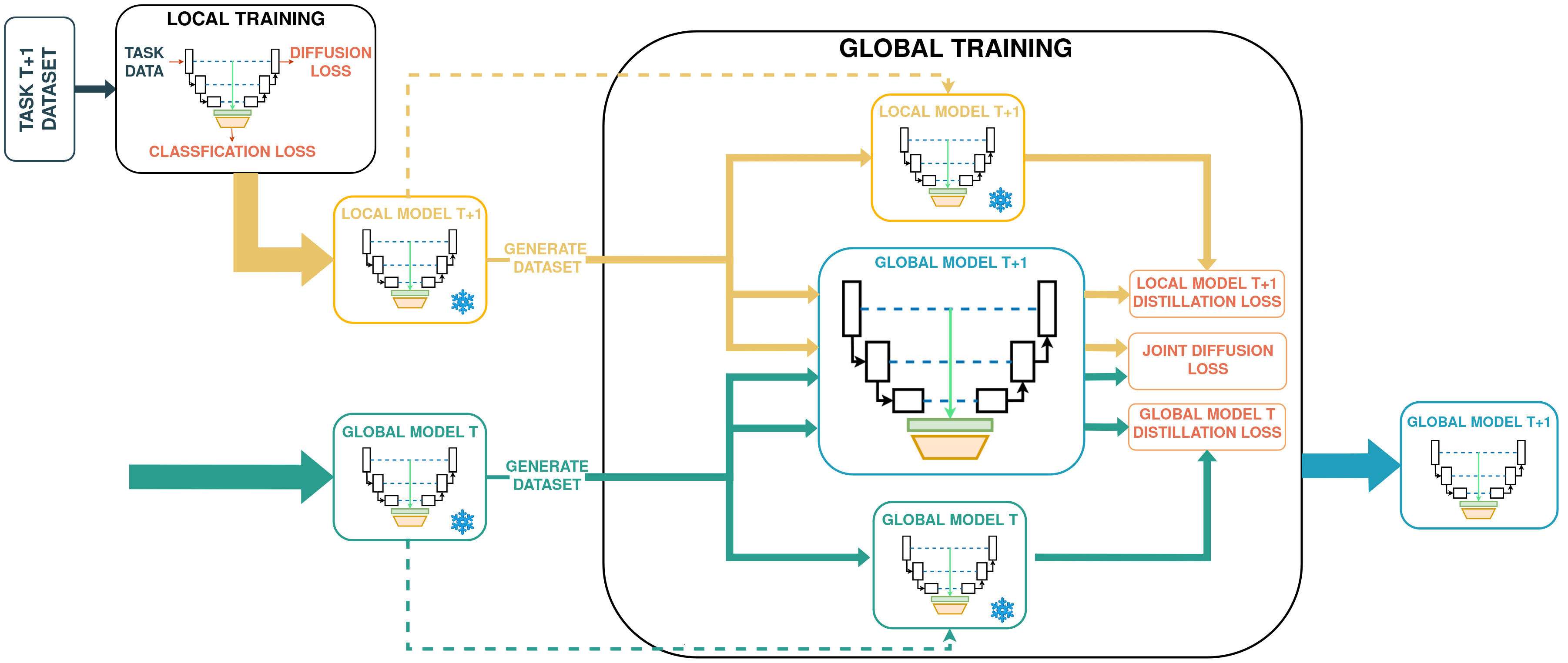}
            \vskip -1mm
            \caption{Overview of \ours{}. To adapt the global model to task $\tau+1$, we first train a local joint diffusion model on task $\tau+1$ only. We then use the global model and the local model to generate a synthetic dataset consisting of samples from tasks $1, 2 \dots \tau+1$. Finally, we fine-tune the global model with the generated dataset and a combination of joint diffusion and knowledge distillation losses.}
            \vspace{-0.4cm}
            \label{fig:cl_method}
        \end{figure*}

        Following the introduction in the previous section, we introduce \ours{} - a new continual learning method with generative replay that allows for stable adaption to new tasks.
        Our approach takes advantage of three main ideas: joint-diffusion parametrization, two-stage local-to-global training, and knowledge distillation.
        \vspace{-0.4cm}
        \paragraph{Joint diffusion modeling in CL}
        The main component of our solution is based on the joint-diffusion model. We adapt the method described in section Sec.~\ref{sec:jd} to the generative replay strategy. We train jointly a single model with generative and discriminative objectives, and benefit from this fact, by using the generative properties of our model to generate rehearsal samples. 
        
        \vspace{-0.4cm}
        \paragraph{Two-stage local-to-global training}
        One of the crucial problems in continually trained neural networks is the plasticity-stability dilemma. As presented in Fig~\ref{fig:motivation_experiment}, we can easily achieve high stability of our continually trained model through a self-rehearsal mechanism combined with knowledge distillation. However, to improve the plasticity, we relate to the two-stage training~\cite{deja2022multiband, gomez2024plasticity} where we first fine-tune the copy of the current model using only newly available data, which results in perfect plasticity. Then, we distill the adapted model with the previous one into a new global model as presented in Fig.~\ref{fig:cl_method}.
        
        More precisely, we first create a copy of our current joint diffusion model $p_{\nu, \psi, \omega}(x_{0:T}, y)$, and train it with the currently available dataset $D_{\tau}$. We call that model \textit{local} and we optimize it with the loss provided in Equation~\ref{eq:final_joint}.

        Following that, we move to the global stage. We begin by generating a rehearsal dataset $S_{1 \dots \tau-1}$ for tasks $1$ to $\tau-1$ by sampling from the original \textit{global} joint diffusion model. We then combine this dataset with a \textbf{synthetic dataset $S_{\tau}$}, which we generate with the \textit{local} model. All the samples are generated unconditionally by the diffusion part of the joint diffusion. The labels for those generations are then assigned by the classifier part of the appropriate models.

        Finally, to acquire the next \textit{global} model, we fine-tune the previous \textit{global} joint diffusion on the combined synthetic dataset with a joint diffusion loss $L_{JD}$ defined in Equation~\ref{eq:final_joint}.
        \vspace{-0.4cm}
        \paragraph{Knowledge distillation}
         To further mitigate catastrophic forgetting and improve the model's stability, on top of the standard joint diffusion loss, we apply the knowledge distillation technique. To that end, we adapt standard KD loss to our joint model, by summing up separate parts from the diffusion and classifier. For a frozen joint diffusion model $p^f$, updated model $p_{\nu, \psi, \omega}(x_{0:T}, y)$ and dataset $D$ we define the diffusion knowledge distillation loss as:
        \begin{equation}\label{eq:l_t_simple_kd}
            L^{KD}_{t, \text{diff}}(\nu, \psi) = \mathbb{E}_{\mathbf{x}_{0} \in D, \boldsymbol{\epsilon_f}}\left[\left\|\boldsymbol{\epsilon_f}-\boldsymbol{\hat{\epsilon}}\right\|^{2}\right],
        \end{equation}
        where $\boldsymbol{\epsilon_f}$ is the noise predicted by the frozen model $p^f$. 
        
        Classification knowledge distillation loss $L^{KD}_\text{class}(\nu, \omega)$ is then a cross-entropy loss between the vectors of probabilities $\varphi^f$ and $\varphi$ assigned for the same input by models $p^f$, and $p_{\nu, \psi, \omega}$ respectively:
        \begin{equation}
            L^{KD}_\text{class}(\nu, \omega) = -\mathbb{E}_{\mathbf{x}_0} \left[ \sum_{k=0}^{K-1} \log \frac{\exp \left( \varphi_{k} \right)}{\sum_{c=0}^{K-1} \exp \left( \varphi_{c} \right)}  \varphi^f_k \right],
        \end{equation}
        Therefore, the full joint diffusion knowledge distillation loss is defined as a sum:
        \begin{align}
            L_{JDKD}(\nu, \psi, \omega; p^f) = & \alpha_{KD} \cdot L^{KD}_\text{class}(\nu, \omega) - L^{KD}_{0}(\nu, \psi)\notag\\
            & - \sum_{t=2}^{T} L^{KD}_{t, \text{diff}}(\nu, \psi) - L^{KD}_T(\nu, \psi), 
        \label{eq:kd_joint}
        \end{align}
        where $\alpha_{KD}$ is a knowledge distillation classification loss scale. 
        
        Finally, the training objective, which we optimize to obtain the \textit{global} model for task $t$ can be expressed as:
        \begin{align}
        L_{CL}(\nu, \psi, \omega) = & \mathbb{E}_{\mathbf{x}_0, y \in S_{1 \dots t-1}} \left[L_{JDKD}(\nu, \psi, \omega; p^f)\right] \nonumber \\
        & + \mathbb{E}_{\mathbf{x}_0, y \in S_t} \left[L_{JDKD}(\nu, \psi, \omega; p_n)\right] \nonumber \\
        & + \beta \cdot \mathbb{E}_{\mathbf{x}_0, y \in S_{1 \dots t}} \left[L_{JD}(\nu, \psi, \omega)\right] \notag, 
        \end{align}
        where $p^f$ is the previous frozen \textit{global} joint diffusion model, $p_n$ is the frozen \textit{local} model, $S_{1 \dots t}$ is the synthetic dataset for tasks $1$ to $t$ and $\beta$ is a non-knowledge-distillation loss scale.

\vspace{-0.4cm}
\paragraph{Joint Diffusion Model in semi-supervised setup}
        One of the biggest advantages of our approach is its flexibility. The design of \ours{} specifically allows for stable consolidation of knowledge from two fully independent models. That means that during the \textit{local} training stage -- the only phase that deals with the data from the current task -- we can train the \textit{local} model without any consideration for the incremental objective. This observation allows us to easily adapt our method to the semi-supervised continual setup.
        In this section, we describe changes in the \textit{local} training stage that allow us to continually train the model on partially labeled data.
    
        First of all, thanks to our joint modeling, we can benefit from our shared parametrization and simply use the unlabeled examples only with the generative part of our model. However, to further improve the results, we propose to combine our joint modeling with consistency regularization and pseudo-label techniques introduced in \cite{fixmatch}. To that end, similarly to~\cite{fixmatch} 
        we use two different sets of augmentations. With weak augmentations $\alpha$, we assign pseudo-labels to the unlabeled examples, and then, with strong augmentations $A$, we calculate additional training objective 
         $L_\text{ssl}(\nu, \omega)$.
        To obtain an artificial label $q_b$ for unlabeled data $x_\text{u}$ 
        we calculate the argmax of our model's prediction for weakly augmented image $\varphi_q$: $\varphi_q = g_{\omega}(e_\nu(\alpha(x_\text{u}))); \hat{q_b} = \text{arg max}(\varphi_q)$. Then, we  calculate the semi-supervised classification loss $l_\text{ssl}$ as a cross-entropy loss between the model's output for a strongly-augmented data and the assigned pseudo-label: 
        \begin{align}\label{eq:obj_ssl}
            L_\text{ssl}(\nu, \omega) = & -\mathbb{E}_{\mathbf{x}_0, y} \{\mathds{1}[max(\varphi_q) \ge \tau] \notag \\ 
            & H(\hat{q_b}, g_{\omega}(e_\nu(A(x_\text{u}))))\}
        \end{align}
        where $\tau$ corresponds to the threshold below which we discard a pseudo-label and $H(p, q)$ denotes the cross-entropy between two probability distributions $p$ and $q$. 

        The full loss function in our semi-supervised approach is then the following:
        \begin{align}\label{eq:final_joint_ssl}
            L(\nu, \psi, \omega) = &\alpha \cdot L_\text{class}(\nu, \omega) + \alpha \cdot L_\text{ssl}(\nu, \omega) - L_0(\nu, \psi) \notag \\ & - \sum_{t=2}^{T} L_{t, \text{diff}}(\nu, \psi) - L_T(\nu, \psi) . \notag
        \end{align}
    
        Training the model in a semi-supervised setup, over a batch of data $(x_0, y, u_0)$ proceeds very similarly to the training in a fully supervised setup. Firstly, we follow the training steps from the supervised setup, with one exception - instead of noising and then calculating the denoising model's training loss on $x_0$ we use both $x_0$ and $u_0$. Following that, we compute pseudo-labels for unlabeled data and use them to calculate the semi-supervised loss.

\section{Experiments}

\subsection{Experimental Setup}

\paragraph{Datasets} We evaluate our method on three commonly used datasets -- CIFAR10, CIFAR100~\cite{krizhevsky2009learning} and ImageNet100~\cite{deng2009imagenet}. The first two of those datasets contain 50000 training and 10000 test images of size $32 \times 32$ divided into 10 or 100 classes. ImageNet100 on the other hand contains 120000 train images in $64 \times 64$ resolution from 100 different classes. 
For the class-incremental setup, we split the CIFAR10 and ImageNet100 datasets into 5 equal tasks and the CIFAR100 dataset into 5 or 10 equal tasks. For semi-supervised setups, we train our models on two different ratios of labeled data in the dataset, i.e. 0.8\% and 5\%. Those ratios correspond to 4 and 25 labeled examples per class in CIFAR100, and to 40 and 250 samples for CIFAR10.

\begin{table}[t]
\caption{Comparison of our \ours{} with non-generative distillation-based techniques and generative rehearsal methods, including feature replay techniques that we mark in gray color. Results of competing methods from~\cite{cywiński2024guide,szatkowski2024adapt}.}
\label{tab:main_results}
\centering
\begin{small}
\begin{sc}
\resizebox{\linewidth}{!}{ 
\begin{tabular}{lcccc}
\toprule
\multirow{2}{*}{\textbf{Method}} & \multicolumn{1}{c}{CIFAR-10} & \multicolumn{2}{c}{CIFAR-100} & \multicolumn{1}{c}{ImageNet100} \\
 & $T=5$ & $T=5$ & $T=10$ & $T=5$ \\
\midrule
Joint & 93.14 $\pm$ 0.16 & \multicolumn{2}{c}{72.32 $\pm$ 0.24} & 66.85 $\pm$ 2.25 \\
Continual Joint & 86.41 $\pm$ 0.32 & 73.07 $\pm$ 0.01  & 64.15 $\pm$ 0.98 & 50.59 $\pm$ 0.35 \\
Fine-tuning & 18.95 $\pm$ 0.20 & 16.92 $\pm$ 0.03 & 9.12 $\pm$ 0.04 & 13.49 $\pm$ 0.18 \\
\midrule
EWC & 20.08 $\pm$ 0.51 & 22.28 $\pm$ 1.03 & 13.95 $\pm$ 1.01 & 21.35 $\pm$ 0.26 \\
GKD (LWF) & 49.60 $\pm$ 0.42& 37.46 $\pm$ 1.18  & 27.91 $\pm$ 0.47 & 33.90 $\pm$ 0.25\\
MKD (iCaRL) & 50.37 $\pm$ 2.15 & 34.01 $\pm$ 0.69 & 25.95 $\pm$ 0.50 &33.03 $\pm$ 0.23\\
TKD (SS-IL) & 50.46 $\pm$ 1.96 & 38.47 $\pm$ 0.83 & 28.96 $\pm$ 0.19 & 35.22 $\pm$ 0.22\\
ANCL & 44.70 $\pm$ 3.24 & 29.46 $\pm$ 0.28 & 18.76 $\pm$ 0.95 &29.40 $\pm$ 0.23\\
\midrule
DGR VAE  & 28.23 $\pm$ 3.84 & 19.66 $\pm$ 0.27 &10.04 $\pm$ 0.17 & 9.54 $\pm$ 0.26 \\
DGR+distill & 27.83 $\pm$ 1.20 & 21.38 $\pm$ 0.61  & 13.94 $\pm$ 0.13 & 11.77 $\pm$ 0.47 \\
RTF  & 30.36 $\pm$ 1.40 & 17.45 $\pm$ 0.28 & 12.80 $\pm$ 0.78 & 8.03 $\pm$ 0.05 \\
MeRGAN  & 51.65 $\pm$ 0.40 & \phantom09.65 $\pm$ 0.14  & 12.34 $\pm$ 0.15  & - \\
\rowcolor{lightergray} BIR  & 36.41 $\pm$ 0.82 & 21.75 $\pm$ 0.08 & 15.26 $\pm$ 0.49 & 8.63 $\pm$ 0.19 \\
\rowcolor{lightergray} GFR  & 26.70 $\pm$ 1.90 & 34.80 $\pm$ 0.26 & 21.90 $\pm$ 0.14 & 32.95 $\pm$ 0.35 \\
DDGR & 43.69 $\pm$ 2.60 & 28.11 $\pm$ 2.58 & 15.99 $\pm$ 1.08 & 25.59 $\pm$ 2.29 \\
DGR diffusion & 59.00 $\pm$ 0.57 & 28.25 $\pm$ 0.22 & 15.90 $\pm$ 1.01 & 23.92 $\pm$ 0.92 \\

GUIDE & 64.47 $\pm$ 0.45 & 41.66 $\pm$ 0.40 & 26.13 $\pm$ 0.29 & 39.07 $\pm$ 1.37  \\
\ours{} & \textbf{83.69} $\pm$ 1.44 & \textbf{47.95} $\pm$ 0.61 & \textbf{29.04} $\pm$ 0.41 & \textbf{54.53} $\pm$ 2.15 \\

\bottomrule
\end{tabular}
} 
\end{sc}
\end{small}
\end{table}

\vspace{-0.4cm}
\paragraph{Implementation details}  We use the same joint diffusion model architecture in all experiments on CIFAR10 and CIFAR100, and an upscaled version of it for ImageNet100 experiments to adjust to the bigger dataset resolution. For the UNet, we follow the implementation provided in~\cite{dhariwal2021diffusion}. For the classifier, we use a 2-layer feed-forward net with a hidden layer of 1024 and a LeakyRELU activation function. The detailed description of the training hyperparameters is provided in the Appendix, and in the code repository\footnote{https://github.com/pskiers/Joint-Diffusion-in-Latent-Space}. 

\subsection{Main results} 
\label{sec:main_results}

To compare our method with other generative rehearsal approaches, we measure its performance on several CL benchmarks. In Tab.~\ref{tab:main_results}, we show that our method outperforms other evaluated generative replay techniques in terms of average accuracy after the last task. 
On CIFAR10 and ImageNet100, our method approaches the soft upper limit, calculated as a continual joint method trained with an infinitely long replay buffer. On those benchmarks, we surpass the previous state-of-the-art score by over 19 and 15 points, which is an increase of around 30\% and 40\% for CIFAR10 and ImageNet100 respectively.

\begin{figure}
    \centering
    \includegraphics[width=\linewidth]{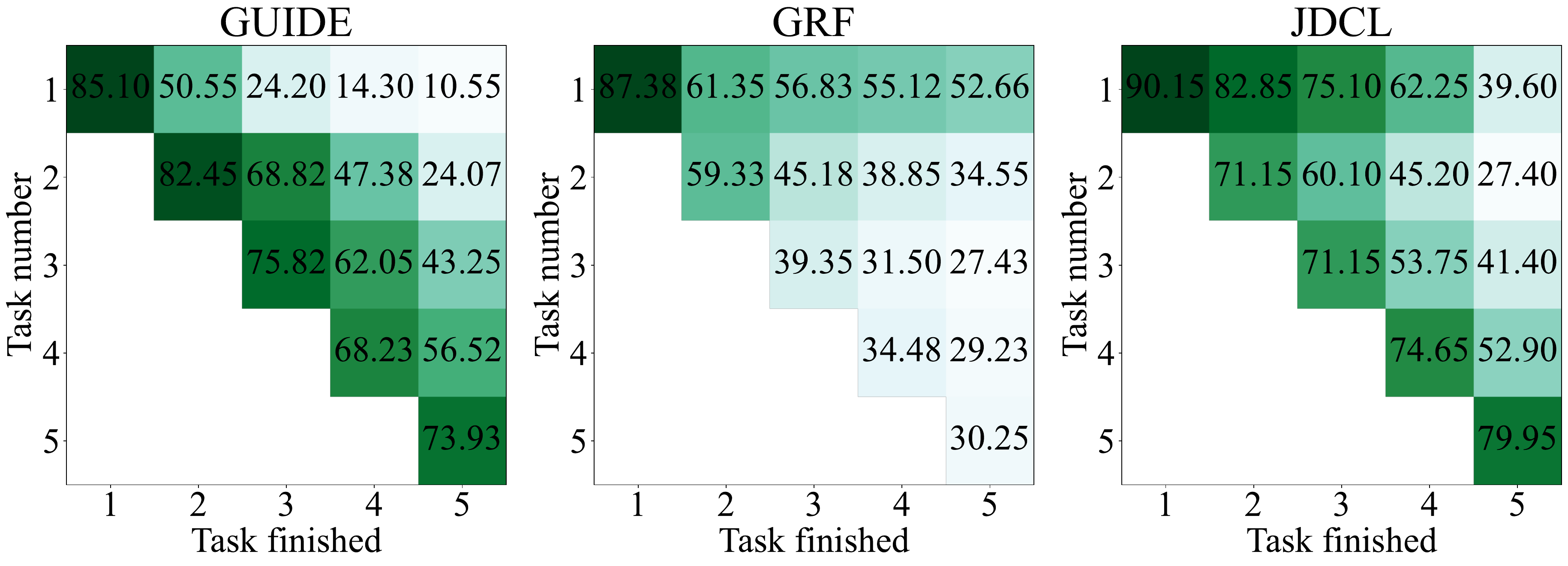}
    \caption{Accuracy on each task after each phase of incremental training on CIFAR100 with 5 tasks.}
    \label{fig:accuracy_cifar100}      
    \vskip -1em
\end{figure}

To better understand the advantages of \ours{} over the competing methods, we perform a detailed comparison of our method against the GFR and GUIDE. To that end, in Fig.~\ref{fig:accuracy_cifar100} we present the accuracy of each approach, on each task after each training phase on CIFAR100 with 5 tasks. \ours{} significantly outperforms GUIDE in terms of knowledge retention from preceding tasks. This is especially visible when we examine the change in performance of the methods on the older tasks. Whereas GUIDE's knowledge obtained during the first task is quickly forgotten and is almost completely lost at the end of the training, \ours{} is able to maintain a good accuracy on the first task for the whole training process. Notably, the performance degradation in our method seems to be related rather to the increasing complexity of the problem instead of forgetting, as we do not see fast degradation on the oldest tasks.
On the other hand, when comparing to the GFR, we can observe that our method, though worse at preserving knowledge, is much better at assimilating new information. This phenomenon can be attributed to the significantly limited plasticity of the GFR that slows down the training of its feature extractor and restricts its update to the feature distillation loss. Consequently, the model has a substantially worse final performance.

\begin{table}[t]
  \centering
    \small
    \caption{Average accuracy with a standard deviation of different methods tested with 5-task CIFAR-10 and 10-task CIFAR-100 semi-supervised settings. The number in brackets indicates the size of the memory buffer for the labeled data. Results of competing approaches from~\cite{kang2022soft}
  }
  \resizebox{\linewidth}{!}{
  \begin{tabular}{l*4c}
  \toprule

    
    \multirow{2}[1]{*}{\textbf{Method}} & \multicolumn{2}{c}{\textbf{CIFAR-10}} & \multicolumn{2}{c}{\textbf{CIFAR-100}}   \\
\cmidrule(lr){2-5}
     & 0.8\% labeled & 5\% labeled  & 0.8\% labeled & 5\% labeled \\
    \midrule
    Fine-tuning & 13.6$\pm$2.9 & 18.2$\pm$0.4 &  1.8$\pm$0.2 & 5.0$\pm$0.3  \\
    \midrule 
    ER (500)          & 36.3$\pm$1.1 & 51.9$\pm$4.5  & 8.2$\pm$0.1 & 13.7$\pm$0.6 \\
    iCaRL (500)      & 24.7$\pm$2.3 & 35.8$\pm$3.2 &  3.6$\pm$0.1 & 11.3$\pm$0.3\\
    FOSTER (500)       & 43.3$\pm$0.7 & 51.9$\pm$1.3 &  4.7$\pm$0.6  & 14.1$\pm$0.6 \\
    X-DER (500)      &  33.4$\pm$1.2 & 48.2$\pm$1.7 & 8.9$\pm$0.3  & 18.3$\pm$0.5 \\
    \midrule 
    PseudoER (500)    & 50.5$\pm$0.1& 56.5$\pm$0.6 &    8.7$\pm$0.4   & 11.4$\pm$0.5 \\
    CCIC (500)        & 54.0$\pm$0.2 & 63.3$\pm$1.9 &             11.5$\pm$0.7& 19.5$\pm$0.2\\
    PAWS (500)        & 51.8$\pm$1.6 & 64.6$\pm$0.6 & 16.1$\pm$0.4 & 21.2$\pm$0.4 \\
    CSL (500)        & 64.5$\pm$0.7  & 69.6$\pm$0.5  & 23.6$\pm$0.3 & 26.2$\pm$0.5 \\
     NNCSL (500) & 73.2$\pm$0.1&  77.2$\pm$0.2  & \textbf{27.4}$\pm$0.5 & \textbf{31.4}$\pm$0.4  \\
    \midrule
    PseudoER (5120)         &  55.4$\pm$0.5 & 70.0$\pm$0.3 &   15.1$\pm$0.2   & 24.9$\pm$0.5\\

    CCIC (5120)        & 55.2$\pm$1.4 & 74.3$\pm$1.7  &              12.0$\pm$0.3& 29.5$\pm$0.4\\
    ORDisCo (12500)   & 41.7$\pm$1.2  & 59.9 $\pm$1.4  & - & - \\
        CSL (5120)        & 64.3$\pm$0.7 & 73.1$\pm$0.3 &   23.7$\pm$0.5         & 41.8$\pm$0.4\\
    NNCSL (5120)    & 73.7$\pm$0.4 &  79.3$\pm$0.3 &  \textbf{27.5}$\pm$0.7 & \textbf{46.0}$\pm$0.2 \\
    \midrule
    \ours{} & \textbf{78.93}$\pm$0.72 & \textbf{79.96}$\pm$0.86 & 22.19$\pm$0.3 & 26.39$\pm$1.7 \\
    \bottomrule
  \end{tabular}
  }

  \label{tab:cifar}
\end{table}
\vspace{-0.4cm}
\paragraph{Semi-supervised continual learning} 
Apart from the standard supervised setup, we evaluate our method on a set of semi-supervised continual scenarios and compare it to the existing approaches. As shown in Tab.~\ref{tab:cifar} \ours{} outperforms other methods on all CIFAR10 setups, while also achieving comparable accuracy on CIFAR100. It is worth noting that \ours{} is able to reach state-of-the-art performance on CIFAR10 without a memory buffer, while also in a majority of tasks sampling fewer examples from the diffusion model than the buffer size in the related works. For CIFAR100, we believe that the reduced performance of our method is caused by an extremely weak signal from the classifier, which introduces an imbalance in our joint training. To further explain the issue, we visualize how \ours{} builds latent representations in the SSL setup in the appendix.

\subsection{Additional Experiments}
        
\begin{table}
\small
  \caption{Ablation study on CIFAR10/5 depicting the importance of individual components. Our method greatly relies on the synergy of joint modeling with knowledge distillation. 
  }
  \centering
  \begin{tabular}{ccc|c}
  \toprule
    \makecell{Joint \\ modeling}& \makecell{Knowledge \\ distillation} & \makecell{2-stage \\ training} & Accuracy\\


    \midrule 
    \ding{51}&\ding{51}&\ding{51}&83.7\\
    \ding{55}&\ding{51}&\ding{51}&68.4\\
    \ding{51}&\ding{55}&\ding{51}&48.2\\
    \ding{51}&\ding{51}&\ding{55}&36.7\\
    \ding{55}&\ding{55}&\ding{51}&32.7\\
    \bottomrule

  \end{tabular}

  \label{tab:ablation}
\end{table}

\paragraph{The interplay between shared parametrization, distillation, and two-stage training}~\newline
In this work, we propose a new generative replay method for continual learning. Apart from the main idea of our approach, which is joint diffusion and classification modeling, we combine it with two important techniques: knowledge distillation and two-stage training. In this section, we ablate the impact of each component. Notably, for experiments without joint modeling, we use a separate diffusion model with the same architecture as the joint diffusion model, and ResNet-18 as a classifier. We train and distill them in the same manner as the joint model. In Tab.~\ref{tab:ablation} we show the results for the model trained without individual components.
As visible, high performance of \ours{} is attributed to the synergy of the proposed joint modeling, knowledge distillation technique, and two-stage training. 
To further highlight the 
interplay between those three components, we perform additional experiments. 

\begin{figure}[t]
    \centering
        \centering
        \footnotesize
        \includegraphics[height=2.8cm]{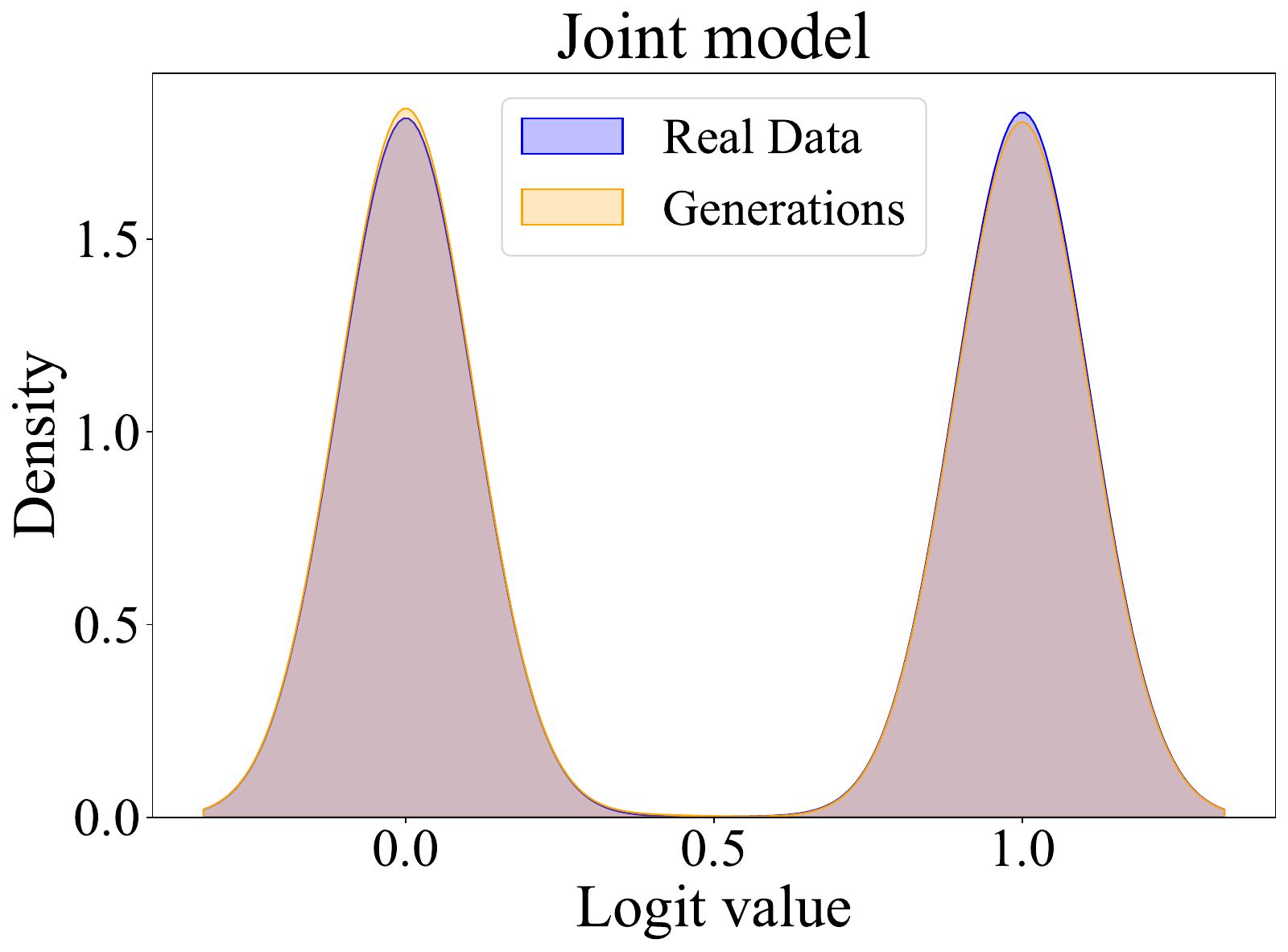}
        \label{fig:acc_abl_1}
        \includegraphics[height=2.8cm]{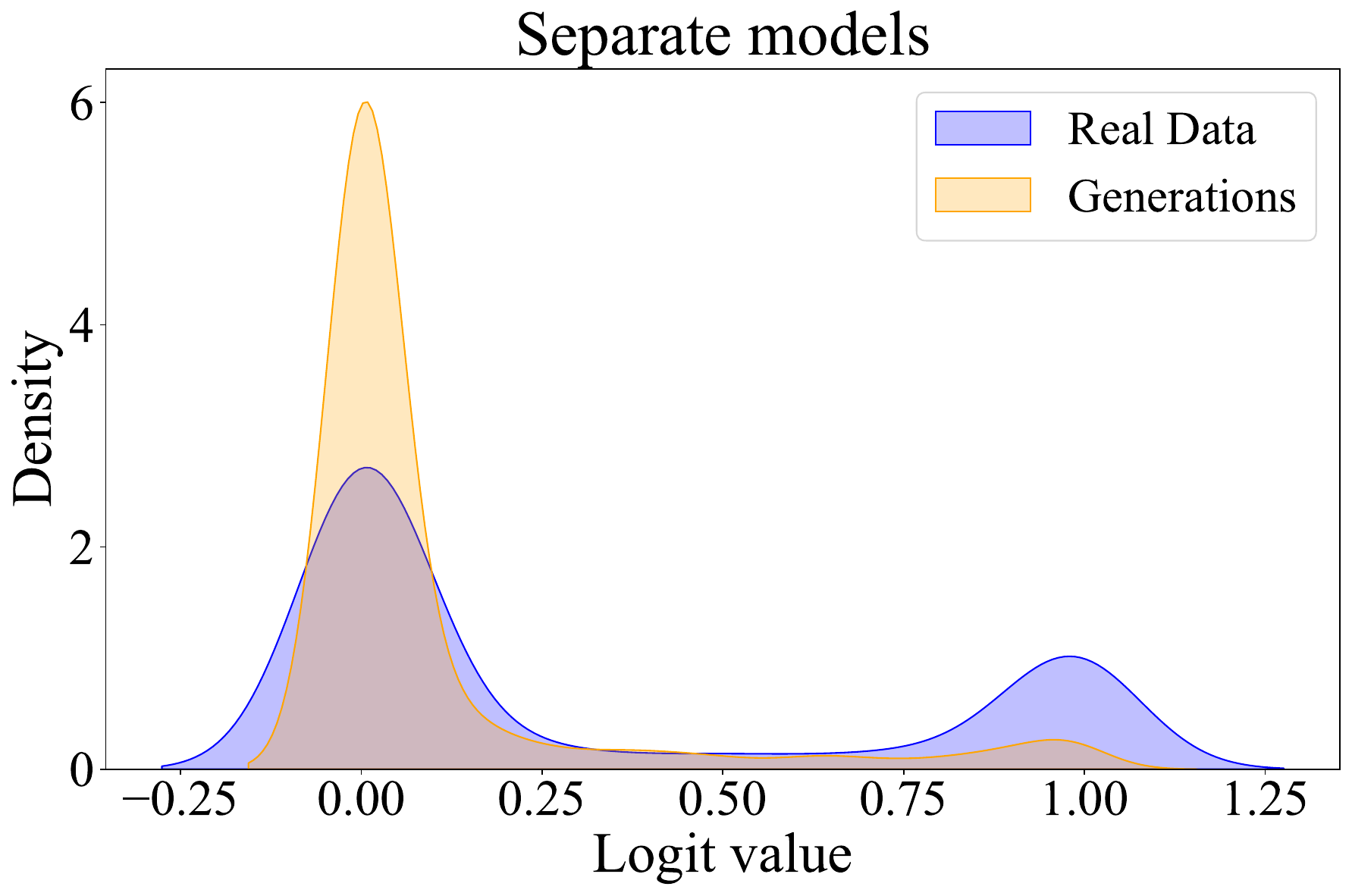}
    \vskip  -0.3cm
    \caption{Difference between the distribution of logit value corresponding to class 1 from CIFAR-10 for real and synthetic data. Samples generated by the joint model closely match the real data's logits distribution, whereas those from the separately trained classifier diverge significantly.}
    \label{fig:logits}
\end{figure}

First, we emphasize the importance of joint diffusion for effective knowledge distillation. To that end, we train a joint diffusion model and a separate classifier and diffusion on a single task. 
In Fig.~\ref{fig:logits}, we compare the classifier's logits distributions for a single class from the real data and rehearsal samples generated by the diffusion model trained either separately or jointly with the classifier. As visible, generations from the joint model closely match the real data logits distribution. However, for the separate classifier, the distributions diverge significantly. This discrepancy means that when updating the global model via knowledge distillation, the rehearsal data remains in-distribution for the joint classifier while being OOD for the separate classifier, which explains the observed performance gap. 

\begin{figure}[t]
    \centering
    \begin{subfigure}{0.49\linewidth}
        \centering
        \includegraphics[width=\linewidth]{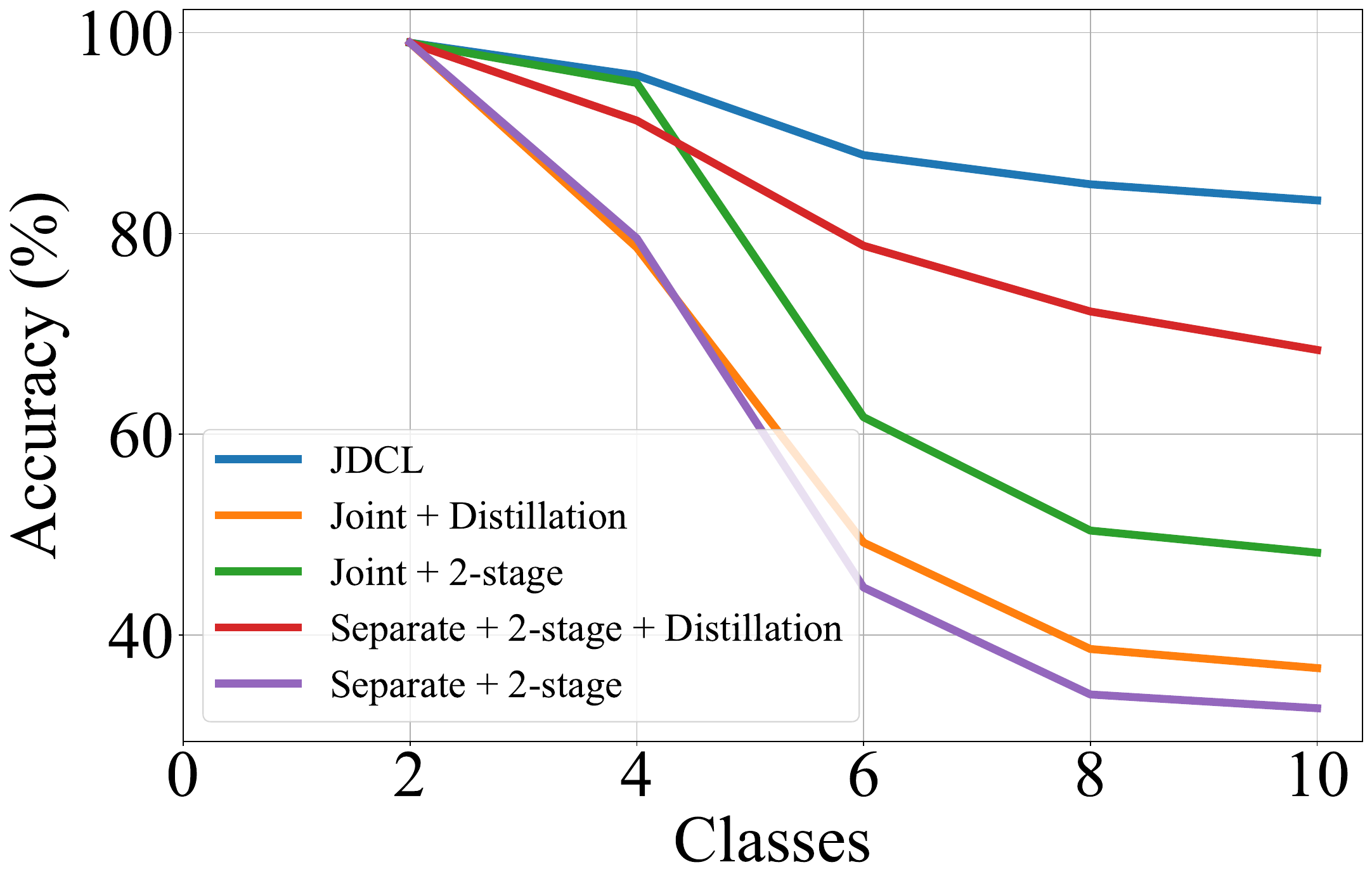}
        \label{fig:ablation_acc_tasks}
    \end{subfigure}
    \hfill
    \begin{subfigure}{0.49\linewidth}
        \centering
        \includegraphics[width=\linewidth]{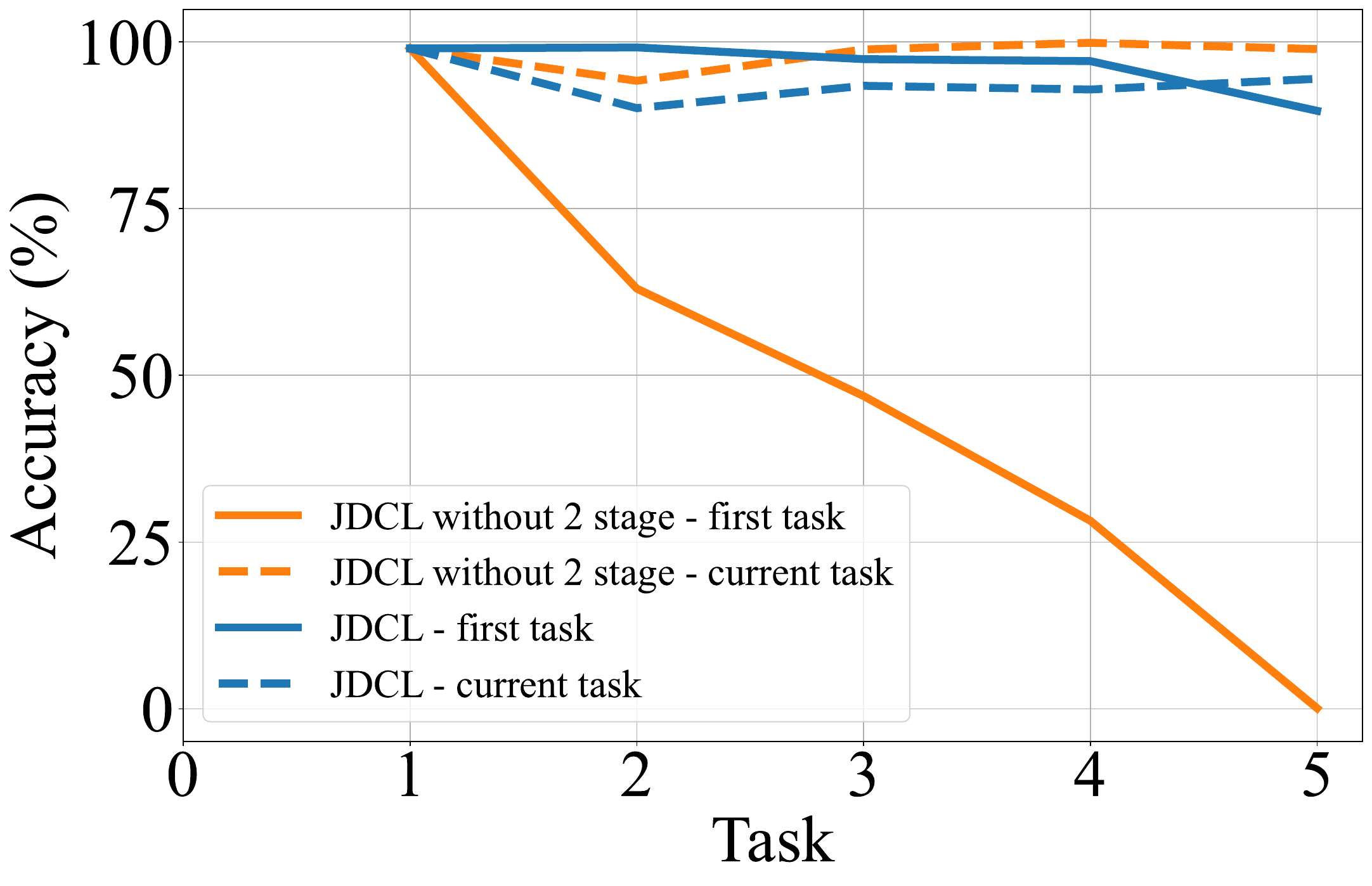}
        \label{fig:2stage_ablation}
    \end{subfigure}
    \vskip -0.4cm
    \caption{Mean accuracy after each task for the ablation methods (left) and accuracy of \ours{} on current and the first task with and without two-stage training (right)}
    \label{fig:ablation_comparison}
    \vskip -0.4cm
\end{figure}

To further highlight this synergy, in Fig~\ref{fig:ablation_comparison} (left) we present how the average accuracy changes after each task for the ablated models. We observe that knowledge distillation and two-stage training can largely mitigate the absence of joint modeling on a task-to-task basis. However, in the CL setup, the performance differences steadily accumulate, leading to significantly worse final results. 

Finally, we show that the use of two-stage training with distillation significantly stabilizes the alignment of new and old knowledge. In Fig.~\ref{fig:ablation_comparison} (right) we plot the accuracy on the newly observed task, and the remaining accuracy on the first one. We can observe that although two-stage training cannot achieve as high plasticity, stable distillation from a local copy of a model to the global one leads to significantly reduced forgetting in the first task.

\begin{figure}
    \centering
    \includegraphics[width=\linewidth]{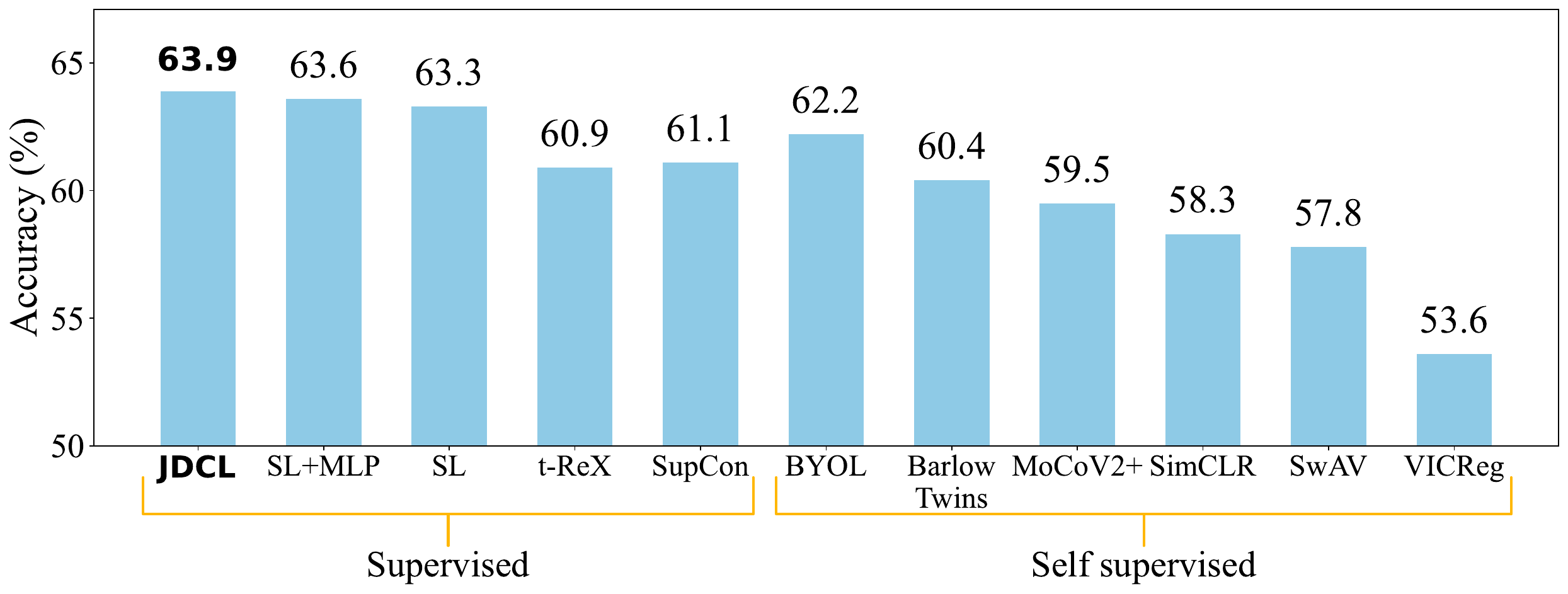}

    \caption{Accuracy of a logistic regressor trained on representations learned with different supervised and self-supervised methods on CIFAR100. Results of competing approaches from~\cite{fini2022self,marczak2025revisiting}}
    \label{fig:self-sup}           
\end{figure}

\vspace{-0.4cm}
\paragraph{\ours{} as a continual representation learner}~\newline
In this section, we show that our method is able to learn meaningful data representation in a continual setup. We follow the evaluation approach common in self-supervised continual learning techniques, as presented in~\cite{fini2022self}. In particular, we report the accuracy of a logistic regressor trained on the features provided by our model i.e. on the set of representations of the UNet model,  pooled into a single vector. In Fig.~\ref{fig:self-sup} we compare our model with supervised and self-supervised continual methods. We can observe that the representations learned by our model are indeed very informative. Moreover, we find that \ours{} outperforms other approaches on the CIFAR100 setup with 5 tasks.

\begin{table}
\centering
\small
  \caption{Comparison of the generation quality of \ours{} and Deep Generative Replay with standard DDPM after the last task.}
  \centering
  \resizebox{\linewidth}{!}{
  \begin{tabular}{l*6c}
  \toprule

    
    \multirow{2}[1]{*}{\textbf{Method}} & \multicolumn{3}{c}{\textbf{CIFAR-10/5}} & \multicolumn{3}{c}{\textbf{CIFAR-100/5}}   \\
\cmidrule(lr){2-7}
     & FID & Precision & Recall & FID & Precision & Recall  \\
    \midrule
    DGR & 49.56 & 0.54 & 0.34 & 50.36 & 0.48 & \textbf{0.30} \\
    JDCL & \textbf{32.12} & \textbf{0.60} & \textbf{0.43} & \textbf{39.93} & \textbf{0.71} & 0.27\\
    \bottomrule
  \end{tabular}
  }
  \vspace{-0.5cm}

  \label{tab:generativ_cl}
\end{table}

\vspace{-0.4cm}
\paragraph{Incremental generative setup comparison}~\newline
Generative-replay-based continual learning methods rely heavily on the underlying generative model to preserve knowledge about the previous tasks. The quality of the samples generated by the method is therefore of utmost importance. In this section, we perform a comparison of \ours{} and the diffusion model trained continually on CIFAR10 and CIFAR100, both split into 5 tasks. We present the achieved FID, Precision, and Recall after the last task in Table \ref{tab:generativ_cl}. \ours{} outperforms the baseline method on all but one metric. We can observe significant improvement in the precision of generations, which results in generated rehearsal samples being clearly assigned to one of the previous classes, bringing less noisy signals to the model. As argued in~\cite{gao2023ddgr}, this trait can be the reason for the improved performance of the continually trained classifier.

\vspace{-0.4cm}
\paragraph{Runtime analysis of \ours{} for CIFAR-10/5}~\newline
One of the biggest drawbacks of generative-replay-based techniques is an excessive computational cost associated with the training of the generative model and additional time spent on sampling of the rehearsal examples. However, thanks to the joint modeling our method is very data-efficient. For example, in the case of the CIFAR dataset, at the beginning of each task, we sample only 400 examples per class for CIFAR-100 or 1000 examples per class for CIFAR10. This occupies approximately 13\% of the total GPU time spent on each task. The combination of the joint model training additionally allows us to achieve state-of-the-art results with computational costs comparable to other diffusion-based solutions. Moreover, with simple tricks such as training with reduced float precision (\ours{} fast) our method achieves performance significantly higher than other approaches, while also having the lowest computational cost. In Table~\ref{tab:runtime_analysis}, we present a runtime analysis to compare the training times of \ours{} with diffusion-based baselines. All methods were trained on the same machine with a single NVIDIA A100 GPU. 

\begin{table}[t]
  \caption{Runtime analysis of all baseline methods on the CIFAR-10/5 benchmark. Our approach is computationally more efficient than related diffusion-based methods.}
  \label{tab:runtime_analysis}
  \centering
  \begin{tabular}{lcc}
    \toprule
    Method & Time [GPU-hours] & Accuracy                  \\
    \midrule
    DDGR & $41.29$ & 43.69  \\
    DGR diffusion   & \underline{28.02}  &  59.00 \\
    GUIDE & 30.60 & 64.47 \\
    \ours{} (fast) & \textbf{27.21} & \underline{78.10} \\
    \ours{} & 53.83 & \textbf{83.69 } \\
    \bottomrule
  \end{tabular}
  \vspace{-0.1cm}
\end{table}

\vspace{-0.5em}
\section{Conclusions}
\vspace{-0.5em}
In this work, we introduce \ours{} -- a new continual learning method that employs joint-diffusion modeling for generative replay. We propose a shared parametrization that mitigates the problem of knowledge transfer in generative replay. In our experimental section, we show that \ours{} outperforms recent generative replay techniques in standard class incremental scenarios. Moreover, we propose the adaptation of our method to semi-supervised learning where we are able to outperform buffer-based approaches. Finally, we show in additional experiments that the performance of our method should be attributed to its ability to learn and maintain useful representations.

\section*{Acknowledgments}
This work is supported by National Centre of Science (NCP, Poland) Grants No. 2022/45/B/ST6/02817, and 2023/51/B/ST6/03004. We acknowledge PLGrid for providing computer facilities under grants no. PLG/2025/018424 and PLG/2025/018551.

{
    \small
    \bibliographystyle{ieeenat_fullname}
    \bibliography{main}
}
\input{sec/X_suppl}

\end{document}

%% file: preamble.tex
\usepackage{xcolor}         
\usepackage{graphicx}
\usepackage{amsmath}
\usepackage{multirow}
\usepackage{colortbl}
\usepackage{wrapfig}
\usepackage{subcaption}
\usepackage{algorithm}
\usepackage{algpseudocode}
\usepackage{pifont}
\usepackage[T1]{fontenc}

\usepackage{makecell}
\usepackage{mathtools}
\usepackage{dsfont}
\usepackage[export]{adjustbox}
\usepackage{wrapfig}
\usepackage{makecell}
\usepackage{float}

\newcommand{\ours}{JDCL}

\definecolor{lightergray}{rgb}{0.9,0.9,0.9}
\definecolor{darkgreen}{rgb}{0., 0.5, 0.}

%% file: sec/X_suppl.tex
\clearpage
\setcounter{page}{1}
\maketitlesupplementary
\appendix

\section{Details on initial experiment - impact of the generative replay}
In this section, we describe the experimental details of the initial experiment presented in Fig~\ref{fig:motivation_experiment}.
For our classifier, we use the ResNet152 architecture with 58.3M parameters, which is comparable to 56.3M parameters of the joint diffusion model. During the experiment, we first train both the classifier and joint diffusion model on the CIFAR10 dataset and save both models. We then use the trained joint diffusion to sample a synthetic dataset - notably, both images and labels are generated by the joint diffusion model. Then, in the first variant of the experiment, we continue training both saved models on this synthetic dataset. In a second variant, we additionally incorporate a knowledge distillation loss during the continued training. For the joint diffusion model, this involves applying the distillation approach detailed in our method, leveraging the saved model trained on the original CIFAR10 data. As for the ResNet152, we utilize the saved classifier trained on the original data and apply only the classification component of the distillation loss of our method.

\section{Main experiments - implementation details}

\label{app:implementation}
We train diffusion using linear noise schedulers over 1000 steps and use DDIM with 250 steps to generate replay samples. In particular, we generate 1000, 400, and 1040 samples per class in CIFAR10, CIFAR100, and ImageNet100 respectively. For the initial local model, we train for 50,000 steps on CIFAR10, 100,000 steps on CIFAR100, and 120,000 steps on ImageNet100. Subsequent tasks are trained for 30,000 steps on CIFAR10, 70,000 steps on CIFAR100, and 90,000 steps on ImageNet100. Additionally, following~\cite{deja2023learning}, for the first 10000 steps of every \textit{global} model training, we do not calculate the classification loss and optimize only the generative objective. We use a single AdamW optimizer with a learning rate set to 0.0002 and a batch size of 256. For all experiments, we set the classification loss scales $\alpha$ and $\alpha_{KD}$ to 0.001, and the $\beta$ hyperparameter that scales joint-diffusion loss to $0.01$.

We define three sets of data augmentations. The first one consists of flipping, resized cropping, and normalization to range $[-1, 1]$, and is used to train the denoising model. The second, \textit{strong} augmentation set includes flipping, cropping, the RandAugment~\cite{cubuk2020randaugment} augmentation policy, and normalization to range $[-1, 1]$. The third and final, \textit{weak} augmentation set consists of flipping, cropping, and normalization to range $[-1, 1]$. We use the \textit{strong} augmentation set to perturb the images for the classifier in the supervised setups. 

Additionally, following~\cite{fixmatch}, we use our \textit{strong} and \textit{weak} augmentation sets to compute the semi-supervised loss. We also set the $\tau$ threshold to $0.95$ and we set the supervised-to-unsupervised batch size ratio to 7, which corresponds to a supervised batch size of 32 and an unsupervised batch size of 224.


\section{Stability of the \ours{}'s representation space in an incremental setup}
\begin{figure}[h]
    \centering
    \includegraphics[width=\linewidth]{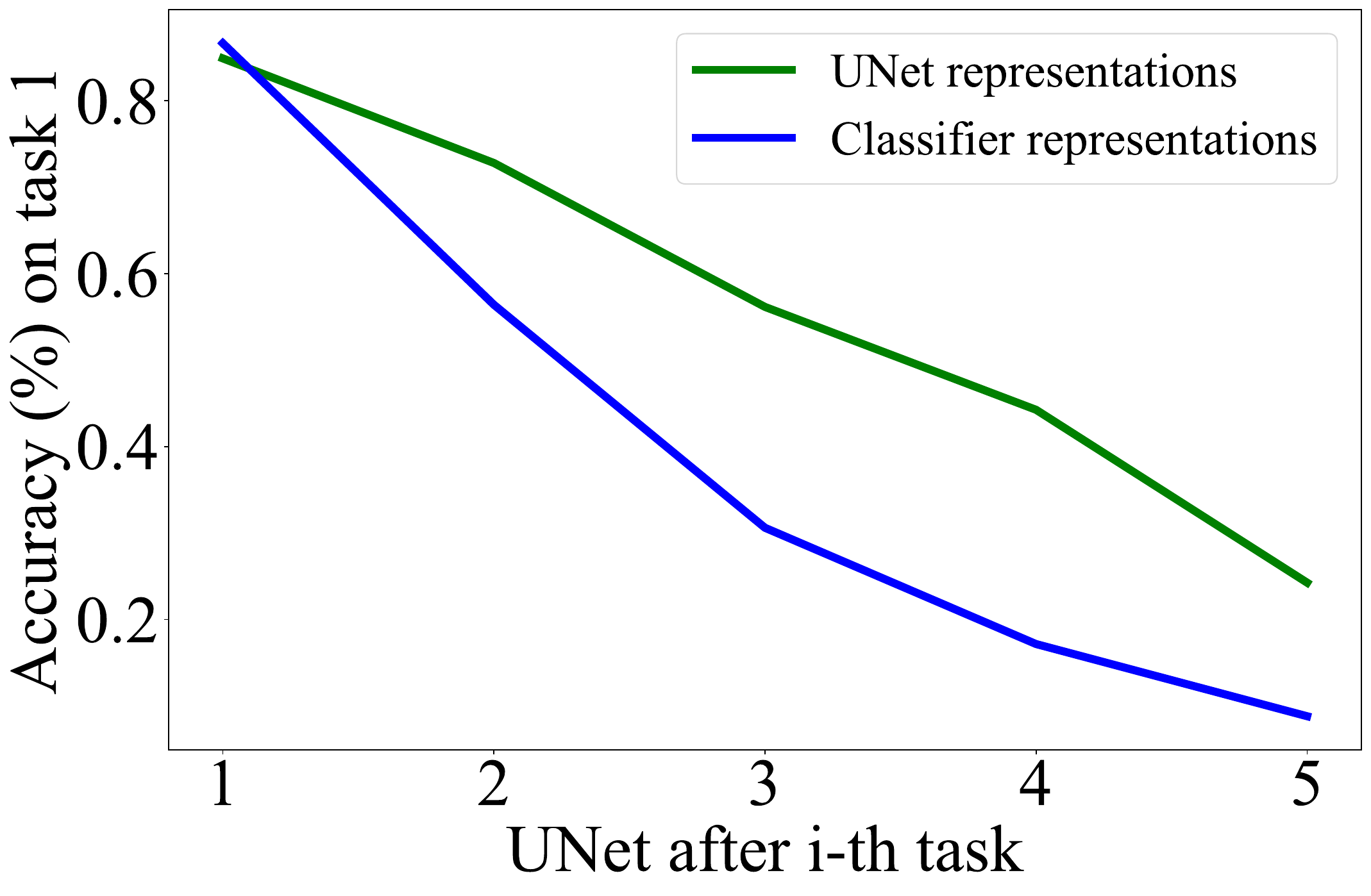}

    \caption{Accuracy of the kNN-clasifier trained on representations provided by the model after the first task of class-incremental CIFAR100 setup with 5 tasks} 
    
    \label{fig:knn}           
\end{figure}

One of the most important features of our methods is its ability to learn a meaningful representation space shared by the generative and discriminative parts of the model. To better understand how this representation space changes during the class-incremental training, we propose to measure the drift of representations, with the following experiment on a 5-task CIFAR100 setup. 
We first extract data representations from two parts of our UNet model, the H-space inside of the generative part of our model, and the penultimate layer of the classifier head.  We train a kNN-classifier on top of those representations provided by the model after the first task. Then, we evaluate how the accuracy of this classifier changes when the representations for data from the first task are extracted with models trained incrementally on later tasks. 
We present the results in Fig.~\ref{fig:knn}. We observe that the representations extracted from the UNet change significantly slower than those from the classifier. This means that representations that are not used for classification drift significantly slower than those that are classifier-specific, which highlights the benefits of joint modeling in a continual setup.

\begin{figure*}[t]
    \vskip -4mm
    \centering
        \includegraphics[width=\textwidth]{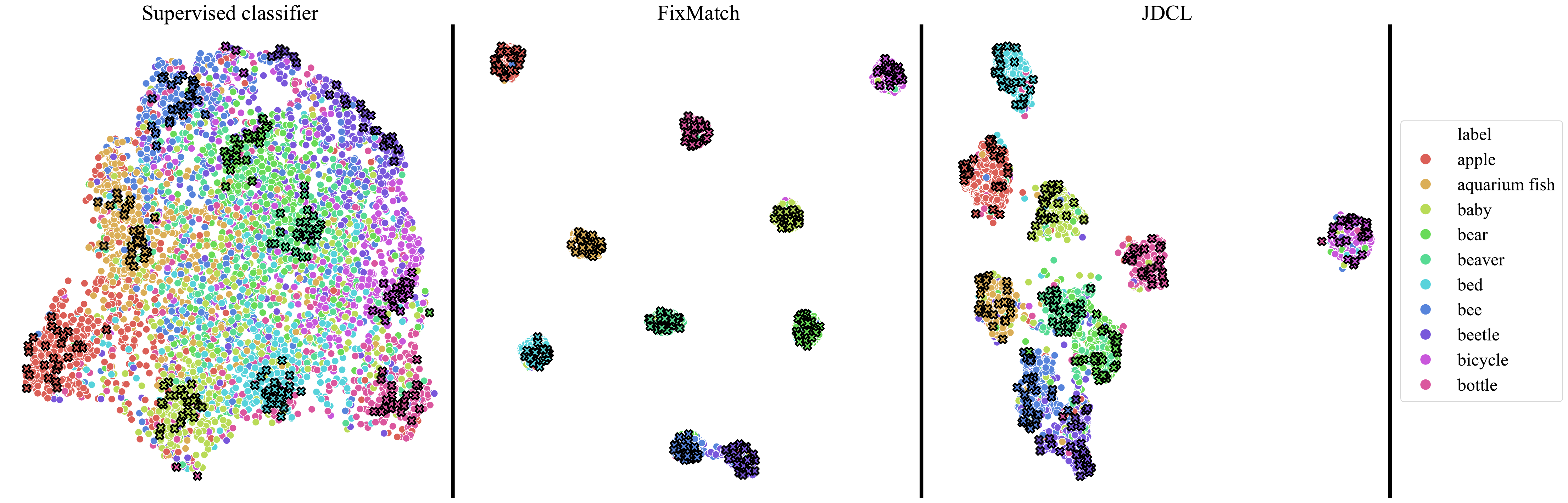}
    \vskip -1mm
    \caption{Visualization of the representation spaces (umap) of different methods trained on a partially labeled dataset. Labeled data samples are marked with an "X", while unlabeled data points are marked with circles. All the models were trained on the first 10 classes of CIFAR100 with 5\% of the data being labeled.}
    \label{fig:cl_in_ssl}
\end{figure*}
\begin{table*}[t]
\caption{Comparison of our \ours{} with generative rehearsal methods including feature replay techniques that we mark in gray color. Results of competing methods from~\cite{cywiński2024guide}.}
\label{tab:avg_forgetting}
\centering
\begin{small}
\begin{sc}
\resizebox{\textwidth}{!}{ 
\begin{tabular}{lcccccccc}
\toprule
 & \multicolumn{4}{c}{Average accuracy $\bar{A}_T$ ($\uparrow$)} & \multicolumn{4}{c}{Average forgetting $\bar{F}_T$ ($\downarrow$)} \\
 \toprule
\multirow{2}{*}{\textbf{Method}} & \multicolumn{1}{c}{CIFAR-10} & \multicolumn{2}{c}{CIFAR-100} & \multicolumn{1}{c}{ImageNet100} & \multicolumn{1}{c}{CIFAR-10} & \multicolumn{2}{c}{CIFAR-100} & \multicolumn{1}{c}{ImageNet100} \\
 & $T=5$ & $T=5$ & $T=10$ & $T=5$ & $T=5$ & $T=5$ & $T=10$ & $T=5$\\
\midrule
Joint & 93.14 $\pm$ 0.16 & \multicolumn{2}{c}{72.32 $\pm$ 0.24} & 66.85 $\pm$ 2.25 & - & - & - & -\\
Continual Joint & 86.41 $\pm$ 0.32 & 73.07 $\pm$ 0.01  & 64.15 $\pm$ 0.98 & 50.59 $\pm$ 0.35 & \phantom02.90 $\pm$ 0.08 & \phantom07.80 $\pm$ 0.55 & \phantom06.67 $\pm$ 0.36 & 12.28 $\pm$ 0.07\\
Fine-tuning & 18.95 $\pm$ 0.20 & 16.92 $\pm$ 0.03 & 9.12 $\pm$ 0.04 & 13.49 $\pm$ 0.18 & 94.65 $\pm$ 0.17 & 80.75 $\pm$ 0.22 & 87.67 $\pm$ 0.07 & 64.93 $\pm$ 0.00 \\
\midrule
DGR VAE  & 28.23 $\pm$ 3.84 & 19.66 $\pm$ 0.27 &10.04 $\pm$ 0.17 & 9.54 $\pm$ 0.26 & 57.21 $\pm$ 9.82& 42.10 $\pm$ 1.40 & 60.31 $\pm$ 4.80 & 40.46 $\pm$ 0.91\\
DGR+distill & 27.83 $\pm$ 1.20 & 21.38 $\pm$ 0.61  & 13.94 $\pm$ 0.13 & 11.77 $\pm$ 0.47 &  43.43 $\pm$ 2.60 & 29.30 $\pm$ 0.40 &21.15 $\pm$ 1.30 & 41.17 $\pm$ 0.43 \\
RTF  & 30.36 $\pm$ 1.40 & 17.45 $\pm$ 0.28 & 12.80 $\pm$ 0.78 & 8.03 $\pm$ 0.05 & 51.77 $\pm$ 1.00& 47.68 $\pm$ 0.80 & 45.21 $\pm$ 5.80 & 41.2 $\pm$ 0.20\\
MeRGAN  & 51.65 $\pm$ 0.40 & \phantom09.65 $\pm$ 0.14  & 12.34 $\pm$ 0.15  & - & - & - & - & - \\
\rowcolor{lightergray} BIR  & 36.41 $\pm$ 0.82 & 21.75 $\pm$ 0.08 & 15.26 $\pm$ 0.49 & 8.63 $\pm$ 0.19 & 65.28 $\pm$ 1.27 & 48.38 $\pm$ 0.44 & 53.08 $\pm$ 0.75 & 40.99 $\pm$ 0.36 \\
\rowcolor{lightergray} GFR  & 26.70 $\pm$ 1.90 & 34.80 $\pm$ 0.26 & 21.90 $\pm$ 0.14 & 32.95 $\pm$ 0.35 & 49.29 $\pm$ 6.03 & \textbf{19.16} $\pm$ 0.55 & \textbf{17.44} $\pm$ 2.20 & 20.37 $\pm$ 1.47 \\
DDGR & 43.69 $\pm$ 2.60 & 28.11 $\pm$ 2.58 & 15.99 $\pm$ 1.08 & 25.59 $\pm$ 2.29 & 62.51 $\pm$ 3.84 & 60.62 $\pm$ 2.13 & 74.70 $\pm$ 1.79 & 49.52 $\pm$ 2.52\\
DGR diffusion & 59.00 $\pm$ 0.57 & 28.25 $\pm$ 0.22 & 15.90 $\pm$ 1.01 & 23.92 $\pm$ 0.92 & 40.38 $\pm$ 0.32 & 68.70 $\pm$ 0.65 & 80.38 $\pm$ 1.34 & 54.44 $\pm$ 0.14 \\

GUIDE & 64.47 $\pm$ 0.45 & 41.66 $\pm$ 0.40 & 26.13 $\pm$ 0.29 & 39.07 $\pm$ 1.37 & 24.84 $\pm$ 0.05 & 44.30 $\pm$ 1.10 & 60.54 $\pm$ 0.82 & 27.60 $\pm$ 3.28 \\
\ours{} & \textbf{83.69} $\pm$ 1.44 & \textbf{47.95} $\pm$ 0.61 & \textbf{29.04} $\pm$ 0.41 & \textbf{54.53} $\pm$ 2.15 & \textbf{11.13} $\pm$ 3.27 & 37.20 $\pm$ 0.75 & 59.55 $\pm$ 0.66 & \textbf{19.51} $\pm$ 6.27\\

\bottomrule
\end{tabular}
} 
\end{sc}
\end{small}
\end{table*}

\section{Representation space in a semi-supervised setup}
In this work, we introduce a new method of training joint diffusion model in a semi-supervised setup. In this section, we show that this approach allows our model to learn shared representations that are meaningful for both generative and discriminative objectives. To that end, we perform a visual comparison of the representation spaces in Fig.~\ref{fig:cl_in_ssl}. We compare our method with a state-of-the-art semi-supervised method FixMatch~\cite{fixmatch}, and a baseline classifier model that we train only on the labeled data. To extract the representations, we average pool the UNet representations into a single vector in our approach, while for the other two models, we take the output of the second to last layer. We observe that our approach is able to group the labeled and unlabeled data from the same class into similar regions of the embedding space shared between the classifier and diffusion model, similar to the state-of-the-art FixMatch approach.

\begin{figure*}[t]
    \centering
    \includegraphics[width=\linewidth]{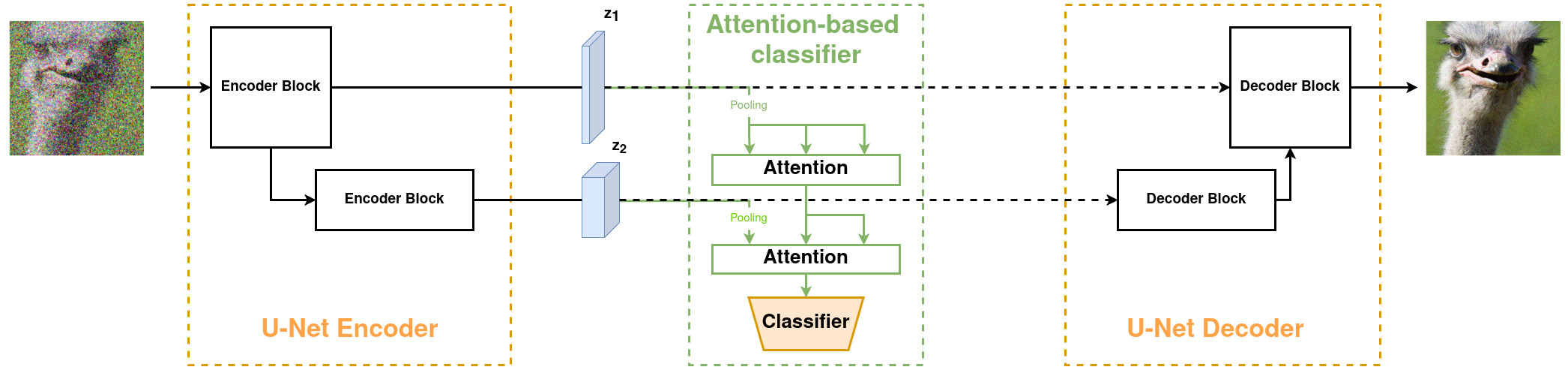}
    \label{fig:att_cls}
    \vspace{-10pt}
    \caption{Visualization of the Joint Diffusion with attention-based classifier. The classification method processes the representations from the shallowest to the deepest levels of UNet architecture using the attention mechanism to create a feature vector.}

\end{figure*}
\begin{table*}[!h]
        \centering
        \begin{sc}
        \caption{Comparison of the classification accuracy and FID scores of the Joint Diffusion model trained with the attention-based and with the pool-based classifier.}
        \smallskip
        \resizebox{\linewidth}{!}{
        \def\arraystretch{1.1}
        \begin{tabular}{l||ccc|ccc}
            \hline
             \multirow{2}{*}{\textbf{Method}} & \multicolumn{3}{c|}{CIFAR-10} &   \multicolumn{3}{c}{CIFAR-100} \\
            \cline{2-7}
            & \multicolumn{1}{c|}{all labels} & \multicolumn{1}{c|}{1000 labels} & \multicolumn{1}{c|}{250 labels} & \multicolumn{1}{c|}{all labels} & \multicolumn{1}{c|}{10000 labels} & \multicolumn{1}{c}{2500 labels}\\
            \midrule
            \textbf{JD: Attention - acc} & \multicolumn{1}{c|}{95.7\%} & \multicolumn{1}{c|}{\textbf{93.9\%}} & \multicolumn{1}{c|}{\textbf{90.9\%}} & \multicolumn{1}{c|}{\textbf{78.3\%}} & \multicolumn{1}{c|}{\textbf{72.9\%}} & \textbf{63.4\%} \\
            \textbf{JD: Pooling - acc} & \multicolumn{1}{c|}{\textbf{96.4\%}} & \multicolumn{1}{c|}{92.1\%} & 86.6\% & \multicolumn{1}{c|}{77.6\%} & \multicolumn{1}{c|}{69.9\%} & 60.8\%\\
            \midrule
            \textbf{JD: Attention - FID} & \multicolumn{1}{c|}{15.9} & \multicolumn{1}{c|}{28.2} & \multicolumn{1}{c|}{28.5} & \multicolumn{1}{c|}{23.1} & \multicolumn{1}{c|}{-} & - \\
            \textbf{JD: Pooling - FID} & \multicolumn{1}{c|}{7.9} & \multicolumn{1}{c|}{\textbf{27.0}} & \multicolumn{1}{c|}{\textbf{27.4}} & \multicolumn{1}{c|}{\textbf{17.4}} & \multicolumn{1}{c|}{-} & - \\
            \hline
        \end{tabular}
      }
      \label{tab:att_vs_pool_classification}
    \end{sc}
    \vspace{-0.8em}
\end{table*}

\section{Additional results - average forgetting}
To further evaluate our approach, we use another metric commonly used in continual learning - average forgetting, defined as in~\cite{chaudhry2018riemannian}. In Tab.~\ref{tab:avg_forgetting} we present the results of \ours{} alongside competing generative replay techniques. Our method outperforms all other approaches on both CIFAR10 and ImageNet100. For CIFAR100 benchmarks, although \ours{} outperforms other methods on the average accuracy metric, we observe higher forgetting when compared to the GFR. However, as discussed in Sec.~\ref{sec:main_results} and shown in Fig.~\ref{fig:accuracy_cifar100}, this can be attributed to the limited plasticity of the GFR that significantly restricts its model's update. This approach, while admittedly leading to lower average forgetting, results in worse plasticity, and consequently significantly worse final performance.

\section{Evaluation of generative qualities}
To further evaluate the generative part of our method, we perform additional experiments. In Tab.~\ref{tab:fid_cifar100t5}, we compare the FID scores after every task of the CIFAR100/5 setup. Our method outperforms DGR (and GUIDE) after each task. To better understand the advantage of our approach, in Tab.~\ref{tab:fwt_bwt_fid} we calculate the equivalents of Forward and Backward Transfer (as in DER~\cite{yan2021dynamically} for the FID metric. We observe that our method is significantly more stable and slightly less plastic, 
which is consistent with the results on the discriminative task. Nevertheless, \ours{}'s overall FID performance is much higher, approaching the IID upper bound equal to 29.7.
 \begin{table}[t]
    \caption{Comparison of the FID scores of \ours{} and Deep Generative Replay with standard DDPM after each task of the CIFAR100/5 setup.}
    \label{tab:fid_cifar100t5}
    \centering
    \begin{small}
    \begin{sc}
    \resizebox{\linewidth}{!}{ 
    \begin{tabular}{lccccc}
    \toprule
    \multirow{2}{*}{\textbf{Method}} & \multicolumn{5}{c}{Task} \\
    & 1 & 2 & 3 & 4 & 5 \\
    \midrule
    DGR/GUIDE & 19.08 & 29.52 & 39.11 & 40.75 & 50.36 \\
    JDCL & 18.20 & 28.06 & 29.73 & 31.39 & 39.93 \\
    \bottomrule
    \end{tabular}
    } 
    \end{sc}
    \end{small}
\end{table}

\begin{table}[t]
    \caption{Forward and backward FID transfer of \ours{} and Deep Generative Replay on CIFAR100/5.}
    \label{tab:fwt_bwt_fid}
    \centering
    \begin{small}
    \begin{sc}
    \resizebox{\linewidth}{!}{ 
    \begin{tabular}{lcc}
    \toprule
    \textbf{Method} & Forward Transfer ($\downarrow$) & Backward Transfer ($\downarrow$) \\
    \midrule
    DGR & \textbf{19.80} & 10.22 \\
    JDCL & 22.80 & \textbf{7.80} \\
    \bottomrule
    \end{tabular}
    } 
    \end{sc}
    \end{small}
\end{table}

\section{Other ways to perform classification in the Joint Diffusion model}
We extensively explored the design of the classifier during our preliminary studies. Here, we include a comparison of the selected pooling method with one of those alternative approaches - an attention-based classifier. Instead of pooling features into a single vector, this method applies a chain of attention blocks over the UNet’s h-representations. Each block takes the corresponding layer’s tensor and the previous block’s output to compute inter‑level attention, merging them into a single vector. Consequently, the classifier better exploits the representations and achieves higher accuracy on most tested 
setups, as shown in Tab.~\ref{tab:att_vs_pool_classification}. However, training with the attention-based approach degrades the quality of the generations, making it less suitable for CL.  

\section{Additional Backward and Forward Transfer metrics}
In Tab.~\ref{tab:fwt_bwt}, we present the additional Forward and Backward Transfer metrics for \ours{}, GUIDE, and GRF on CIFAR100/5 setup. The calculated metrics support our claims from Fig.~\ref{fig:accuracy_cifar100}. \ours{} significantly outperforms GUIDE on Backward and GRF on Forward Transfer.

\begin{table}[t]
    \caption{Comparison of Forward and Backward Transfer of \ours{}, GRF and GUIDE on CIFAR100/5 setup.}
    \label{tab:fwt_bwt}
    \centering
    \begin{small}
    \begin{sc}
    \resizebox{\linewidth}{!}{ 
     \begin{tabular}{lcc}
        \toprule
        \textbf{Method} & Forward Transfer ($\uparrow$) & Backward Transfer ($\uparrow$) \\
        \midrule
        GUIDE & \textbf{-15.72} & -26.86 \\
        GRF & -49.99 & \textbf{-10.00} \\
        JDCL & \underline{-16.61} & \underline{-14.83} \\
        \bottomrule
    \end{tabular}
    } 
    \end{sc}
    \end{small}
\end{table}

\section{Details on semi-supervised scenario}
All experiments follow offline CL protocols, while our semi-supervised scenario builds on the NNCSL~\cite{kang2022soft}, designed specifically for SSL in CL. In the paper authors adapt various offline approaches to the SSL framework, and we adhere to their evaluation protocol. Due to an additional diffusion model, we extend JDCL training to 60k batch updates for CIFAR10.

\section{Training complexity - discussion}
While joint training increases the complexity of a single DM training step, it eliminates the need to train a separate classifier. Consequently, our joint model has fewer parameters (56.3 M) than the separate classifier and DM combined (64.7 M). Additionally, because of the synergy of the generative and discriminative parts, our method converges faster than the standard DM. We show that in Tab.~\ref{tab:conv_rate}, where we compare training loss on the first task of CIFAR10/5 of the 2 methods after the same runtime.

\begin{table}[t]
    \caption{Comparison of the training diffusion loss of the jointly and separately trained diffusion model on the first task of CIFAR10/5.}
    \label{tab:conv_rate}
    \centering
    \resizebox{\linewidth}{!}{ 
    \begin{tabular}{lcc}
    \toprule
    \multirow{1}{*}{\textbf{Runtime}} & \multicolumn{1}{c}{Joint diffusion} & \multicolumn{1}{c}{Separate Diffusion} \\
    \midrule
    1h & 0.01785 & 0.01982 \\
    2h & 0.01741 & 0.01844 \\
    3h & 0.01691 & 0.01723 \\
    
    \bottomrule
    \end{tabular}
    } 
\end{table}

\section{Sensitivity of performance with respect to loss terms scales}
For all experiments, we use exactly the same classification loss scales equal to $0.001$ and non-knowledge-distillation loss scale $\beta$ equal to $0.01$. 
However, to present a full picture, 
we present the influence of different $\beta$ values on the final score on CIFAR10/5. Notably, the majority of results are within statistical errors:
\begin{table}[!h]
\vspace{-0.8em}
\centering
\begin{small}
\begin{sc}
\resizebox{\linewidth}{!}{ 
\begin{tabular}{lccc|>{\columncolor{gray!20}}c|cccc}
\toprule
\textbf{$\beta$}  & 0 & 0.001 & 0.005 & \textbf{0.01} & 0.05 & 0.1 & 0.5 & 1\\
\midrule
Accuracy/gain  & -0.59 & -1.37 & -0.37 & \multicolumn{1}{>{\columncolor{gray!20}}c|}{\textbf{83.69$\pm$1.44}} & +0.32 & -1.16 & -6.79 & -10.41 \\
\bottomrule
\end{tabular}
} 
\end{sc}
\end{small}
\end{table}